\documentclass[pdflatex,sn-mathphys-num,iicol]{sn-jnl}%

\usepackage{graphicx}%
\usepackage{multirow}%
\usepackage{amsmath,amssymb,amsfonts}%
\usepackage{amsthm}%
\usepackage{mathrsfs}%
\usepackage[title]{appendix}%
\usepackage{xcolor}%
\usepackage{textcomp}%
\usepackage{manyfoot}%
\usepackage{booktabs}%
\usepackage{algorithm}%
\usepackage{algorithmicx}%
\usepackage{algpseudocode}%
\usepackage{listings}%

\usepackage{makecell}
\usepackage{hyperref}

\theoremstyle{thmstyleone}%

\theoremstyle{thmstyletwo}%

\theoremstyle{thmstylethree}%

\begin{document}

\title[DiffSAC: Diffusion-guided Sampling for Consensus-based Robust Estimation]{DiffSAC: Diffusion-guided Sampling for Consensus-based Robust Estimation}

\author[1,2]{\fnm{Chang} \sur{Nie}}\email{changnie@sjtu.edu.cn}

\author[3]{\fnm{Guangming} \sur{Wang}}\email{gw462@cam.ac.uk}

\author[1,2]{\fnm{Zhe} \sur{Liu}}\email{liuzhesjtu@sjtu.edu.cn}

\author*[1,2]{\fnm{Hesheng} \sur{Wang}}\email{wanghesheng@sjtu.edu.cn}

\affil[1]{\orgdiv{School of Automation and Intelligent Sensing}, \orgname{Shanghai Jiao Tong University}, \orgaddress{\city{Shanghai}, \postcode{200240}, \country{China}}}

\affil[2]{\orgdiv{Key Laboratory of System Control and Information Processing}, \orgname{Ministry of Education of China}, \orgaddress{\city{Shanghai}, \postcode{200240}, \country{China}}}

\affil[3]{\orgdiv{Department of Engineering}, \orgname{Cambridge
University}, \orgaddress{\city{Cambridge}, \postcode{CB2 1TN}, \country{UK}}}

\abstract{
Robust estimation is a core computer vision task frequently tackled using sample consensus. However, traditional methods suffer from inefficient sampling as they struggle to identify effective minimum sets before hypothesis evaluation. To address these challenges, we propose a novel Diffusion-guided Sampling for Consensus-based Robust Estimation (DiffSAC) framework. DiffSAC introduces a diffusion model to learn the distribution of effective minimum sets. It refines the confidence for each data point, indicating whether it belongs to a good minimum set, rather than ranking the data points as in previous work. This significantly reduces the need to process numerous bad sets. To constrain the refinement direction, geometric features are incorporated as conditions within our diffusion model. Consequently, DiffSAC outputs a small number of high-quality minimum sets, enabling identification of the best hypothesis via consensus evaluation. Notably, compared to previous works requiring evaluating over ten thousand hypotheses, DiffSAC achieves state-of-the-art performance with only dozens, significantly boosting efficiency. Extensive experiments across five classic computer vision tasks demonstrate the superiority of DiffSAC. The diffusion model's sampling accelerators enable real-time operation, and DiffSAC can be used as a plug-and-play module to improve existing sample consensus methods.
}

\keywords{Robust Estimation, Diffusion Models, Line Fitting, Fundamental Matrix Estimation, Essential Matrix Estimation}

\maketitle

\section{Introduction}\label{sec1}

Robust estimation is critical to computer vision, underpinning tasks such as simultaneous localization and mapping (SLAM) \cite{wenzel20254seasons, xu2025generalized, zhang2024improved, wu2022yolo, he2024real, zheng2023simultaneous, cheng2024ransac}, motion segmentation \cite{barath2022learning, guo2023feasible, yang2021toward, yang2022ransacs, wei2023generalized, wan2025instance, jiao2021effiscene}, point cloud registration \cite{li2021point, dai2022multisource, chung2024centralized, shi2024ransac, sun2021ransic, qin2022geometric, martinez2022ransac} and structure from motion (SfM) \cite{campos2021orb, zhou2021event, cui2023mcsfm, ghahremani2021direct, won2023robust, zhu2024revisit, yang2021toward}. Despite significant progress, accurately and efficiently estimating models from data contaminated by noise remains a persistent challenge.
To address the challenges of noisy data, sample consensus algorithms, notably RANdom SAmple Consensus (RANSAC) \cite{fischler1981random}, are a popular approach. RANSAC operates by iteratively hypothesizing models from randomly selected minimum sets of data. A minimum set contains the fewest data points required to define a model. For instance, two points suffice for a 2D line. By focusing on minimum sets rather than the entire dataset, RANSAC reduces the influence of outliers. After solving a hypothesis from a minimum set, RANSAC evaluates its consensus, which is quantified by the number of data points consistent with the hypothesis within a defined tolerance. Finding a hypothesis with maximal consensus necessitates repeating this hypothesize-and-test cycle many times.

While RANSAC exhibits robustness and generalizability, its inherent random sampling strategy introduces limitations. Firstly, RANSAC samples minimum sets without prior refinement, leading to considerable variability in their quality. Secondly, the number of iterations required to find a good model grows exponentially with increasing noise levels \cite{chum2008optimal}, significantly impacting efficiency. Thirdly, RANSAC employs uniform random sampling, neglecting potentially valuable geometric information that could guide sampling.

In response to these limitations, many methods focus on improving the sampling process \cite{brachmann2019neural, chum2005matching, magri2017multiple, tiwari2016robust, mateus2023fast, lu2024feature}. Early methods prioritize sampling based on heuristics, such as PROSAC \cite{chum2005matching}, aiming to favor outlier-free minimum sets. Recent neural network-guided methods, like NG-RANSAC \cite{brachmann2019neural}, incorporate probabilistic preferences to enhance sampling. However, these preference-based strategies primarily refine sampling randomness but do not eliminate the risk of selecting bad minimum sets. Furthermore, guiding sampling towards a singular optimal solution can be difficult due to complex data distributions.

For these problems in previous work, the diffusion models \cite{ho2020denoising} demonstrate the potential to solve them. It excels at capturing complex data patterns and generating diverse, high-quality outputs. For example, in text-to-image generation, diffusion models \cite{rombach2022high} can create multiple plausible images from a single text prompt, allowing the selection of the most suitable result. We propose leveraging this generative power of diffusion models to directly produce a small number of reliable and deterministic high-quality minimum sets. Such a deterministic generation strategy can avoid the problem of probabilistic sampling in previous methods that often sample numerous bad minimum sets, thereby enhancing the efficiency and robustness of estimation.

\begin{figure*}[t]
  \centering
   \includegraphics[width=0.99\linewidth]{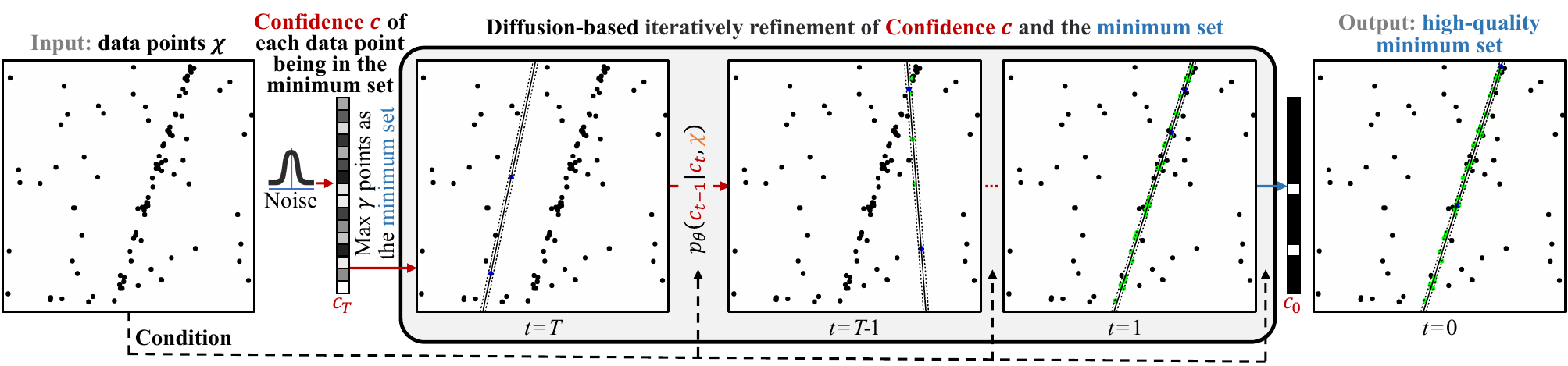}
   \caption{\textbf{Robust estimation with DiffSAC.} The 2D line fitting task is used as an example. DiffSAC predicts the confidence $c$ of each data point being selected as the minimum set with the diffusion model. DiffSAC marries the power of sampling consensus with the strengths of the diffusion model to iteratively refine the minimum set of data points $\chi$. \textbf{\textcolor{green!50!gray}{green}}, \textbf{black} respectively indicate \textbf{\textcolor{green!50!gray} {inliers}}, and \textbf{outliers}.}
   \label{fig:overall}
\end{figure*}

Consequently, we introduce Diffusion-guided Sampling for Consensus-based Robust Estimation (DiffSAC), a novel framework designed to enhance robust estimation tasks. DiffSAC leverages the powerful distribution modeling capacity of diffusion models to improve sample consensus. Specifically, as shown in Fig. \ref{fig:overall}, DiffSAC uses a diffusion model to learn the probability distribution $p(c|\chi)$, which represents assigning a confidence $c$ to each data point in a dataset $\chi$. This confidence $c$ indicates a point belonging to a minimum set. By conditioning the diffusion process on the geometric features of the data $\chi$, the generation of confidence $c$ is guided towards the desired direction. Trained on datasets comprising data points $\chi$ and corresponding optimal minimum sets, DiffSAC can then estimate confidence $c$ for unseen data by sampling from the learned distribution $p(c|\chi)$. This confidence estimation enables DiffSAC to prioritize high-quality minimum sets, substantially improving efficiency by avoiding the exploration of less promising candidates. The inherent stochasticity of diffusion sampling facilitates effective exploration of the probability landscape \cite{ho2020denoising}, making it particularly suitable for confidence prediction. To further enhance model accuracy, DiffSAC generates multiple minimum sets during inference, which are subsequently evaluated using sample consensus to identify the hypothesis with the strongest consensus. This process ultimately selects the best hypothesis from a pool of highly promising candidates.

DiffSAC achieves state-of-the-art performance in five classic robust estimation tasks. Moreover, DiffSAC can be seamlessly integrated into existing robust estimation pipelines as a plug-and-play module. While diffusion models are criticized for computational cost with high-dimensional data like images, the confidence $c$ in DiffSAC is low-dimensional, resulting in minimal overhead. Furthermore, with diffusion model acceleration techniques, DiffSAC can operate in real-time.

The main contributions of DiffSAC are as follows:
\begin{itemize}
	\item We present a novel framework of Diffusion-guided Sampling for Consensus-based Robust Estimation (DiffSAC), which integrates diffusion models into sample consensus for robust estimation, effectively addressing limitations in minimum set refinement and sampling efficiency.
    
	\item DiffSAC leverages geometric features to constrain the diffusion process, iteratively refining confidences to output reliable and deterministic, high-quality minimum sets while minimizing the selection of bad sets, thus improving both efficiency and accuracy.
    
	\item Extensive experiments on 2D line fitting, 3D plane fitting, fundamental matrix estimation, essential matrix estimation, and homography estimation demonstrate that DiffSAC achieves state-of-the-art performance and can be easily integrated into existing sampling consensus methods as a plug-and-play module for diverse robust estimation tasks.
    
\end{itemize}

\begin{figure*}[t]
  \centering
   \includegraphics[width=0.99\linewidth]{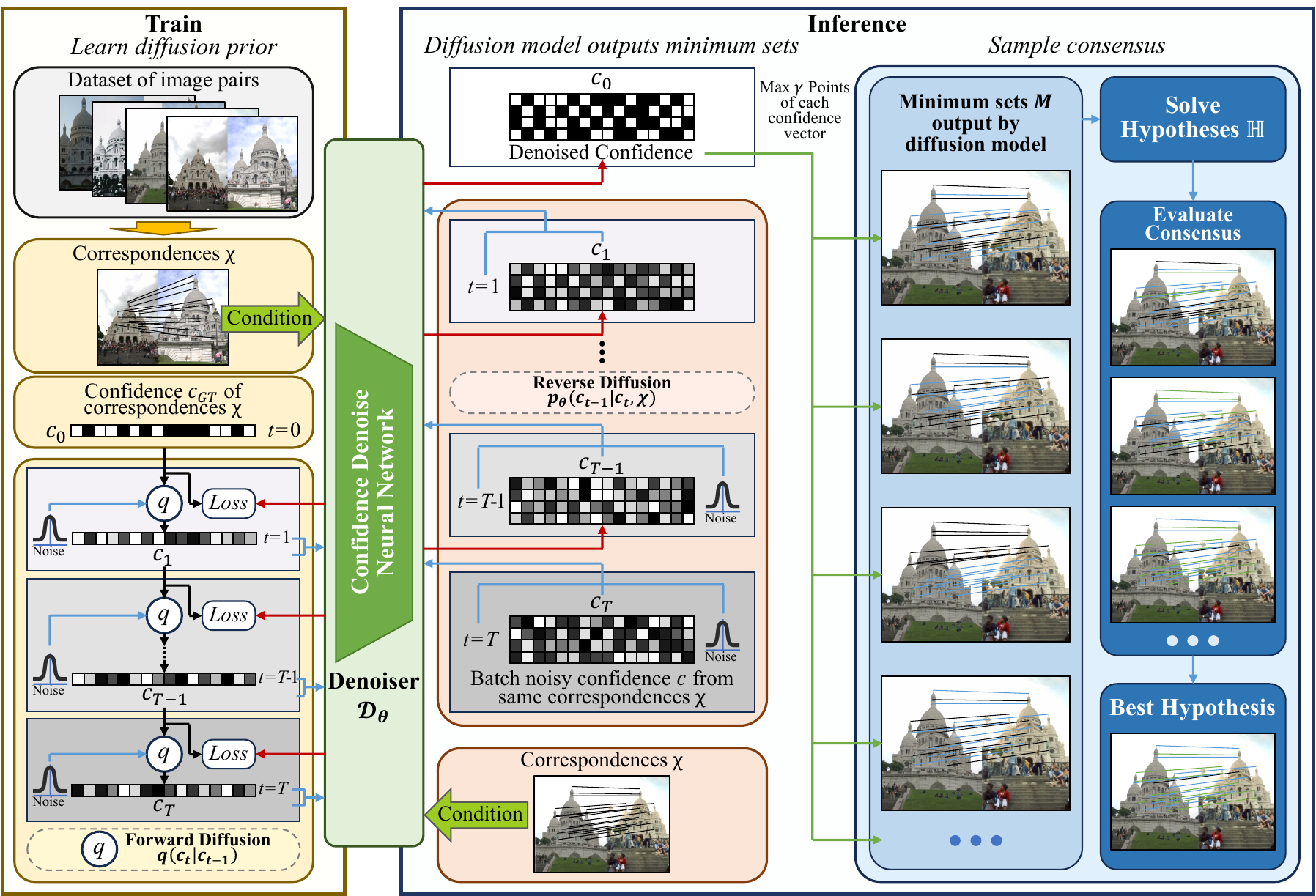}
   \caption{\textbf{The pipeline of the proposed DiffSAC.} We use the fundamental matrix estimation task as an example. DiffSAC uses correspondences $\chi$ as a condition to constrain the generation direction. During training, DiffSAC takes the confidence $c$ of each data point being part of the minimum set to learn the diffusion prior with forward diffusion (Sec. \ref{Prior}). For inference, DiffSAC iteratively refines multiple confidence $c$ simultaneously through reverse diffusion to generate high-quality minimum sets. Then, the hypotheses solved by these minimum sets are evaluated for consensus to obtain the best hypothesis as the final result of robust estimation (Sec. \ref{Diffusion_mini})}

   \label{fig:pipeline}
\end{figure*}

\section{Related Work}
\label{sec:Related_Work}
\textbf{Sample Consensus.}
Robust estimation, essential for handling noisy data, is commonly addressed by sample consensus methods. The Random Sample Consensus (RANSAC) algorithm stands as a foundational approach in this category \cite{fischler1981random}. The strength of RANSAC lies in its simplicity and broad applicability, yet its random sampling process can be inefficient, particularly with high outlier ratios.

To improve the sampling efficiency of RANSAC, subsequent methods have introduced more strategic sampling techniques. PROSAC prioritizes data points more likely to support a valid model by sorting data based on a quality metric and sampling sequentially \cite{chum2005matching}. USAC takes a broader approach, integrating diverse sampling strategies to dynamically balance efficiency and robustness in various scenarios \cite{raguram_usac_2013}. Neural networks are explored to guide sampling. Initial efforts use PointNet to directly classify data points as inliers or outliers \cite{Yi_2018_CVPR}. This approach is further developed to incorporate local context through pooling and unpooling mechanisms, enhancing the classification accuracy \cite{zhang2019learning}. Alternative sampling strategies include NAPSAC, which constrains sampling to local neighborhoods to improve performance in certain cases, though at the risk of overlooking global structure and becoming trapped in local optima \cite{torr2002napsac}. Progressive NAPSAC mitigates this limitation by gradually expanding the sampling area from local to global, seeking a balance between focused and comprehensive search \cite{barath_progressive_2019}. More recently, NG-RANSAC leverages neural networks to predict the probability of selecting effective samples, thus focusing the sampling process on more promising regions of the data space \cite{brachmann2019neural}. DGSAC \cite{tiwari2018dgsac} introduces the concept of Kernel Residual Density (KRD) to create a data-driven, automated pipeline that guides the sampling process and eliminates the need for user-specified parameters like a time budget. However, these methods based on preference sampling still have a higher probability of sampling bad minimum sets. This leads to a lot of unnecessary computing consumption. Utilizing advanced generative methods to directly produce reliable and deterministic high-quality minimum sets can greatly improve efficiency.

In addition to refining the sampling process, many methods focus on improving the quality of generated hypotheses through local refinement. LO-RANSAC refines initial hypotheses by concentrating on the inliers of the current best model, iteratively improving the solution within a local context \cite{chum2003locally}. GC-RANSAC, in contrast, considers the spatial relationships between data points, modeling data as a graph and utilizing graph-cut to achieve a more spatially coherent separation of inliers and outliers \cite{barath2018graph}. MAGSAC++ enhances robustness by employing adaptive thresholds that are less sensitive to variations in noise levels, leading to more reliable estimation in diverse noise conditions \cite{barath_marginalizing_2022}. Deep learning techniques also enhance the sample consensus framework. DSAC replaces the deterministic sampling of RANSAC with a probabilistic approach, allowing for end-to-end training and gradient-based optimization of the sampling process \cite{brachmann_dsac_2017}. Similarly, other methods directly optimize the likelihood of selecting good hypotheses within a deep learning framework, aiming to learn more effective sampling distributions from data \cite{wei_fully_2022}.

\textbf{Diffusion models,} inspired by thermodynamics \cite{sohl2015deep}, learn data distributions through a process of iteratively adding and then removing noise. Diffusion models have demonstrated remarkable success in generating high-quality samples across various data types, including images \cite{li2024snapfusion, zhao2024uni, wu2025paragraph, zhu2025domainstudio, huang2025diffusion}, videos \cite{chen2024videocrafter2, zhou2024upscale, hu2025lamd, xing2024survey, ho2022video}, and 3D point clouds \cite{zheng2024point, kasten2024point, ren2024tiger, zhang2022reggeonet, luo2021diffusion}. Recent work has begun to explore the application of diffusion models in localization and pose estimation. For example, DiffLoc combines diffusion models with pose regression to improve LiDAR localization accuracy \cite{li2024diffloc}, and PoseDiffusion utilizes them to enhance bundle adjustment in structure-from-motion pipelines \cite{wang2023posediffusion}. PC2 leverages diffusion models to reconstruct 3D point clouds from images, guided by camera pose \cite{melas2023pc2}. Despite these advancements, \textit{the potential of diffusion models as a core component within robust estimation frameworks remains largely unexplored.} We believe that the diffusion models are inherently suited to sampling consensus methods and can greatly improve efficiency and accuracy.

\section{Preliminary}
\subsection{Sample Consensus Methods for Robust Estimation} Robust estimation addresses the challenge of determining a reliable hypothesis, denoted as $h$, from a dataset $\chi  = \left\{ {{{\rm{x}}_i}} \right\}_{i = 1}^N$ contaminated with noise. For example, in scenarios involving 2D points, $h$ can represent the parameters of a line fitted to these points. Alternatively, when dealing with image correspondences, $h$ might represent the fundamental matrix in epipolar geometry.

Sample consensus is a prevalent method in robust estimation. This approach operates by first selecting $n$ minimum sets, denoted as $M$, from the dataset $\chi$. The size $\gamma$ of each minimum set is determined by the task. For instance, line estimation in 2D space requires a minimum set size of two points. A solver $\mathcal{S}$ then processes each minimum set to generate a set of candidate hypotheses, $\mathbb{H}$:
\begin{equation}
\mathbb{H}  = \left\{ {\mathcal{S} \left( {{m_j}} \right)\left| {{{m_j}} \in M,j = 1,2,...,n} \right.} \right\}.
\label{solve_h}
\end{equation}
Subsequently, each hypothesis within $\mathbb{H}$ undergoes evaluation using a scoring function $f$ to assess its consensus with the data. A common metric for this evaluation is the inlier ratio, which quantifies the proportion of data points consistent with the hypothesis. The hypothesis achieving the highest score is then chosen as the optimal estimate, $h_{Best}$:
\begin{equation}
{h_{Best}} = \mathop {\arg \max }\limits_{h \in \mathbb{H}} f\left( {h,\chi } \right).
\label{eq:hbest}
\end{equation}
This iterative process of sampling and evaluating multiple minimum sets confers robustness to outliers.

The classical RANSAC algorithm exemplifies this framework, employing random sampling to derive the minimum sets $M$. However, it is recognized that random sampling, particularly without incorporating geometric information, can be inefficient.

\subsection{Diffusion Models}
Diffusion models represent a category of generative models designed to learn intricate data distributions by simulating the reverse of a diffusion process \cite{ho2020denoising, sohl2015deep, song2019generative, song2020denoising, zheng2025diffuvolume, gao2023implicit}. This process introduces noise gradually to data over $T\in \mathbb{N}$ discrete steps, transforming the original data into a noisy state. Training these models involves learning to reverse this noising process, effectively denoising data back to its original structure.

Given a variance schedule $\beta_1, \dots \beta_T$ across $T$ steps, the transition from step $t-1$ to step $t$ during the forward diffusion (noising) process is described as:
\begin{equation}
q( c_ {t} |c_ {t-1}) := \mathcal{N}( c_ {t} ;  \sqrt {1-\beta _ {t}} c_ {t-1} , \beta _ {t} E ),
\label{add_noise}
\end{equation}
where $E$ is the identity matrix. Through this forward process, as $t$ approaches $T$, the data distribution $c_T$ converges to an isotropic Gaussian distribution, $c_T$. By defining $\alpha _t := 1-\beta _t$ and $\overline{\alpha }_t :=  {\textstyle \prod_{s=1}^{t}} \alpha_s$, a closed-form expression allows for direct sampling of $c_{t}$ from $c_0$:
\begin{equation}
c_ {t} \sim  q( c_ {t} |c_ {0})= \mathcal {N} (c_ {t}; \sqrt {\overline{\alpha }_t}c_ {0},(1- \overline {\alpha }_ {t} )E ).
\end{equation}

The reverse diffusion process, $p_ {\theta } ( c_ {t-1} | c_ {t} )$, maintains a Gaussian form if $\beta _ {t}$ values are sufficiently small. This property allows the reverse distribution to be effectively modeled by a denoiser $\mathfrak{D} _ {\theta }$:
\begin{equation}
p_ {\theta } ( c_ {t-1} | c_ {t} ):=\mathcal{N} ( c_ {t-1}; \sqrt {\alpha _ {t} } \mathfrak{D} _ {\theta }(c_ {t},t),(1- \alpha _ {t} )E ).
\label{denoise}
\end{equation}
Iteratively applying this denoising process allows the model to learn to reconstruct the original data distribution from noise, making diffusion models a potent tool for generative tasks.

\section{Methodology}

\subsection{Pipeline of DiffSAC}
DiffSAC leverages diffusion models to enhance sample consensus for robust estimation. The core idea is to use diffusion models to improve both the efficiency of sampling and the quality of the minimum sets. As depicted in Fig. \ref{fig:pipeline}, DiffSAC operates by estimating the conditional probability distribution $p(c|\chi)$ of confidence $c$ for each data point, given the entire dataset. This confidence $c$ represents whether a data point belongs to a good minimum set. DiffSAC approximates this conditional probability distribution using a denoising process inherent to diffusion models.

\begin{figure*}[t]
  \centering
   \includegraphics[width=0.75\linewidth]{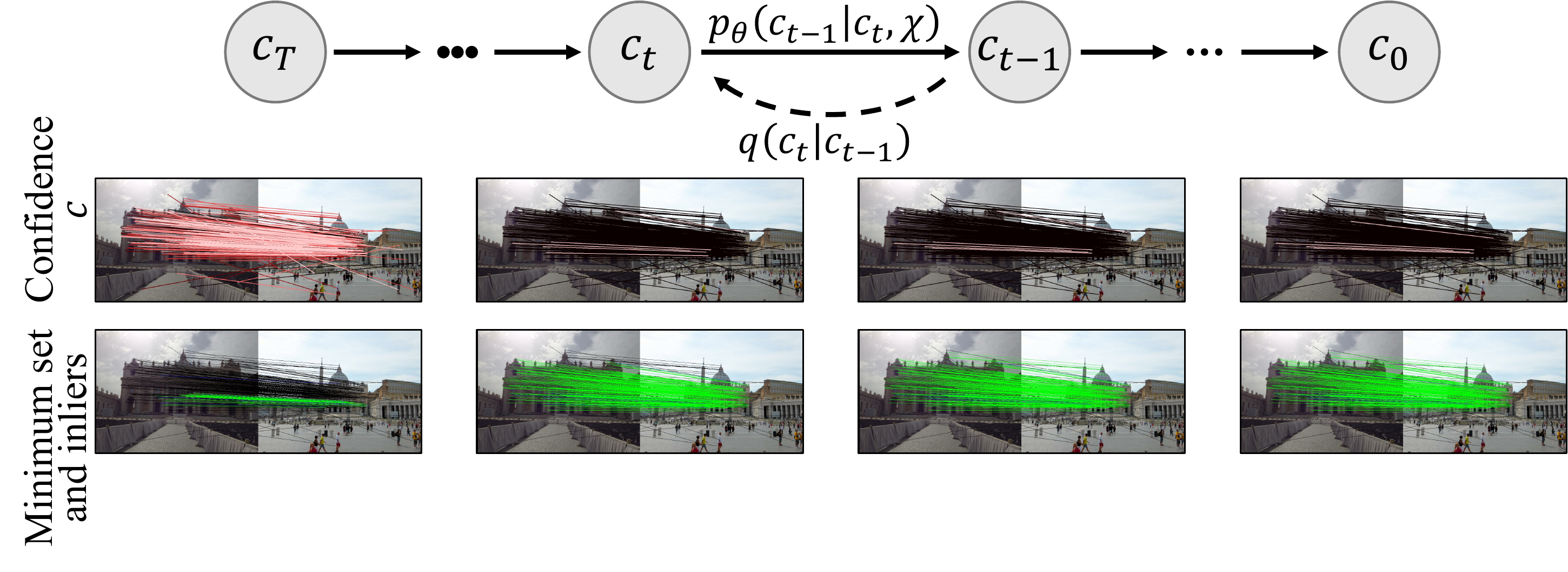}
    \vspace{-0.1cm}
   \caption{\textbf{An illustration of the diffusion process for essential matrix estimation.} The brighter the lines in confidence $c$, the higher the confidence value. Other colors are the same as Fig. \ref{fig:overall}. During forward process, DiffSAC progressively adds Gaussian noise to the ground truth confidence $c_0$. In reverse diffusion, DiffSAC denoises noisy confidence $c_t$ at time $t$ with condition $\chi$.}
   \label{fig:confidence}
\end{figure*}

To achieve this, DiffSAC first trains a diffusion model, denoted as $\mathfrak{D} _ {\theta }$, on a comprehensive dataset $\left \{ \left ( c_k,\chi_k \right ) \right \} _{k=1}^{\mathfrak{S} }$. This dataset comprises pairs of ground truth data points $\chi_k$ and their corresponding confidence $c_k$. During inference, as illustrated in Fig. \ref{fig:confidence}, when presented with a new set of observed data points $\chi$, DiffSAC iteratively refines the sampling from the learned conditional probability distribution $p(c|\chi)$. This iterative refinement estimates the confidence $c$ for each data point, aiming to identify high-quality minimum sets and eliminate bad sets. A key distinction from typical noise generation, which is independent of $\chi$, is that the denoising process of DiffSAC is conditioned on the input data points, making it data-aware. This conditional probability is mathematically expressed as:
\begin{equation}
p_ {\theta } ( c_ {t-1} | c_ {t} , \chi)=\mathcal{N} ( c_ {t-1}; \sqrt {\alpha _ {t} } \mathfrak{D} _ {\theta }(c_ {t},t,\chi),(1- \alpha _ {t} )E ).
\end{equation}

\begin{algorithm}
\caption{Diffusion Training Process}\label{alg:train}
\begin{algorithmic}[1]
\Require  ~~\\
Timesteps $T$;\\
Ground truth confidence $c_0$;\\
Data points $\chi$ as condition;\\
Approximate posterior $q$;\\
Initial denoiser $\mathfrak{D} _ {\theta }$.
\Ensure ~~\\%
Trained denoiser $\mathfrak{D} _ {\theta }$.
\Repeat
\State $c_0\sim q(c_0)$;
\State $t\sim \mathrm{Uniform}({1,\dots ,T})$;
\State Sample confidence $c_t$ at timestamp $t$ through sampling noise $\epsilon \sim \mathcal{N} (0,E)$ in the forward process;
\label{code:fram:extract} %
\State Take gradient descent step on $\nabla_\theta \left \| \mathfrak{D} _ {\theta } ( c_ {t} ,t,\chi )- c_0 \right \|$ ;

\Until convergence;\\
\Return $\mathfrak{D} _ {\theta }$; %
\end{algorithmic}
\end{algorithm}

\subsection{Learn Diffusion Prior}
\label{Prior}
During training, as in Algorithm \ref{alg:train}, the denoiser $\mathfrak{D} _ {\theta }$ is trained to associate input data points $\chi$ with corresponding confidence. This is achieved by constraining the denoiser to generate confidence that aligns with a designated direction learned from the training data. To establish ground truth confidence $c_{GT}$, DiffSAC first determines the optimal minimum set $m_{GT}$ using a ground truth model. Data points within this optimal set are assigned $1$, while the remaining points are assigned $0$. This initial confidence assignment, denoted as $c_0$, is then subjected to a forward diffusion process through Eq. \ref{add_noise} until a predefined time step $t=T$. In each step of forward diffusion, the denoiser $\mathfrak{D} _ {\theta }$ takes noisy confidence $c_t$, the current time step $t$, and the input data points $\chi$ to predict a less noisy confidence $c_{t-1}$. These inputs are unordered data points.

\begin{figure*}[h!]
  \centering
   \includegraphics[width=1.0\linewidth]{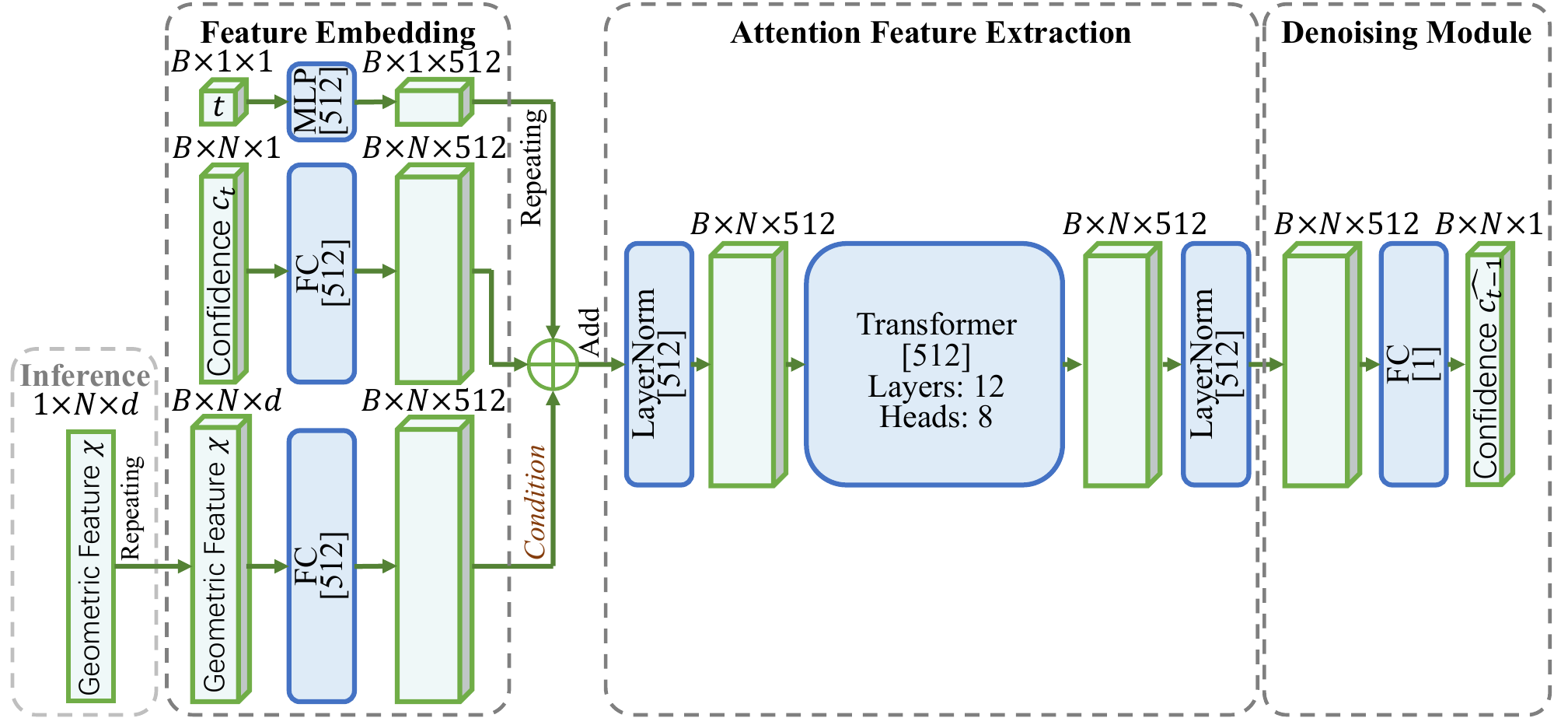}
    \vspace{-0.4cm}
   \caption{\textbf{Architecture of the Confidence Denoise Neural Network in DiffSAC.} The raw data $\chi$ is represented as unordered $N \times d$ data points for conditioning. The network denoises the confidence $c_t$ into $\hat{c}_{t-1}$.}
   \label{fig:denoser}
\end{figure*}

The architecture of the denoiser $\mathfrak{D} _ {\theta }$ is illustrated in Fig. \ref{fig:denoser}. The use of a transformer-based attention mechanism without positional encodings ensures that our denoiser network is permutation-invariant and can naturally handle inputs with a variable number of points (N), as demonstrated in our experiments in Section \ref{plane}. The network $\psi$ consists of three primary modules: Feature Embedding, Attention Feature Extraction, and Denoising Module. In operation, both the noisy confidence $c_t$ and the data points $\chi$ are initially transformed into feature vectors of identical dimensionality using fully connected (FC) layers. The time step $t$ is also embedded into a feature vector using a multi-layer perceptron (MLP). These embedded features are then combined through addition. Subsequently, the combined features are processed by an attention feature extraction module, employing layer normalization followed by a standard transformer network \cite{vaswani2017attention}. The output of the transformer is again layer-normalized. Finally, a fully connected layer maps the processed features to a vector representing the denoised confidence, $\hat{c}_{t-1}$. For the MLP components, $1 \times 1$ convolutions with a stride of 1 are utilized. The operation of the denoiser can be summarized as:
\begin{equation}
\mathfrak{D} _ {\theta } ( c_ {t} ,t,\chi )=\psi  [( c_ {t}^ {i} ,t,\chi^i)_{i=1}^{N} ] = \hat{c}_{t-1},
\end{equation}
where $\psi $ denotes the denoiser taking a sequence of noisy tuples $c_ {t}^ {i}$, diffusion time $t$, and geometric features $\chi^i \in \mathbb{R}^d$. The training process optimizes the diffusion model $\mathfrak{D} _ {\theta }$ by minimizing a denoising loss, calculated as the difference between the predicted denoised confidence $\hat{c}_{t-1}$ and the ground truth confidence $c_0$:
\begin{equation}
\mathcal{L} _ {diff} = E_ {t\sim [1,T] , c_ {t} \sim q( c_ {t}|c_ {0},\chi{} )} [ \left \| \mathfrak{D} _ {\theta } ( c_ {t} ,t,\chi )- c_0 \right \|^ {2}],
\label{loss_diff}
\end{equation}
where the expectation is computed over diffusion timesteps $t$, diffused samples $c_t$, and the training set $\left \{ \left ( c_{0,j},\chi_j \right )  \right \} _{j=1}^{\mathfrak{S} }$.

\begin{algorithm}
\caption{Diffusion Sampling Process}\label{alg:sampling}
\begin{algorithmic}[1]
\Require  ~~\\
Timesteps $T$;\\
Data points $\chi$ as condition;\\
Denoiser $\mathfrak{D} _ {\theta }$.
\Ensure ~~\\%
Refined confidence $c_0$.
\State Sample initial confidence $c_T\sim \mathcal{N} (0,E)$;
\For{$t=T$ to $1$}
\State $z\sim \mathcal{N} (0,E)$ if $t>1$, else $z=0$;
\State $c_{t-1}=\frac{1}{\sqrt{\alpha_t}}\left ( c_t- \frac {1-\alpha _ {t}}{1-\overline {\alpha}_ {t}}\mathfrak{D} _ {\theta } ( c_ {t} ,t,\chi ) \right ) + \sigma _ {t} z
$;
\EndFor \\
\Return $c_0$; %
\end{algorithmic}
\end{algorithm}

\subsection{Inference: Generating and Evaluating Minimum Sets}
\label{Diffusion_mini}
The inference process in DiffSAC is designed to generate and evaluate multiple high-quality minimum sets to find the optimal model hypothesis, as in Algorithm \ref{alg:sampling}. The procedure begins by taking the input data points $\chi$. To efficiently explore the solution space and leverage the stochastic nature of the diffusion model, the input $\chi$ is replicated to form a batch of size $\kappa$. For each of these parallel instances, the process is initiated with a unique, randomly sampled noise vector $c_T\sim N(0,E)$, where $E$ is the identity matrix. This strategy allows the model to generate a diverse pool of candidate solutions simultaneously, which can be efficiently handled using GPU parallelism.

With the batch of $\kappa$ initial noise vectors, the core generation step involves running the reverse diffusion process for $T$ steps. Using the trained denoiser $\mathfrak{D} _ {\theta }$, the model iteratively refines each noise vector, conditioned on the input data $\chi$, to produce $\kappa$ final confidence vectors. At each step $t$ in the reverse sequence $\left ( T,\dots 0 \right )$, the subsequent confidence vector $c_{t-1}$ is sampled from the learned conditional distribution $p_ {\theta } ( c_{t-1}|c_t, \chi )$ as follows:
\begin{equation}
c_{t-1}=\mathcal{N} ( c_ {t-1}; \sqrt {\overline{\alpha} _ {t-1} } \mathfrak{D} _ {\theta }(c_ {t},t,\chi),(1- \overline{\alpha} _ {t-1} )E ).
\end{equation}
This iterative refinement process leverages the learned geometric features of the data to guide the generation, effectively pruning paths that would lead to poor minimum sets and steering the outcome towards high-quality confidence assignments.

Once the $\kappa$ confidence vectors are generated, the next step is to form candidate minimum sets. For each confidence vector $c_0^{(i)}$, a corresponding minimum set is constructed by selecting the $\gamma$ data points from $\chi$ that have the highest confidence values. From each of these $\kappa$ minimum sets, a model hypothesis is solved. This results in a collection of $\kappa$ distinct hypotheses, each derived from a promising minimum set identified by the diffusion model.

Finally, these $\kappa$ hypotheses are evaluated within a sample consensus framework. The consensus score for each hypothesis is calculated by evaluating its consistency with the entire input dataset $\chi$ using an appropriate metric. The hypothesis that garners the highest consensus score is ultimately selected as the final robust estimation result, $h_{Best}$. This comprehensive evaluation ensures that the chosen model is not only derived from a high-quality minimum set but is also the most consistent with the overall data distribution, enhancing the accuracy and robustness of the final estimation.

\section{Experiments}
\label{Experiments}
DiffSAC is evaluated through experiments on five classic tasks: 2D line fitting, 3D plane fitting, fundamental matrix estimation, essential matrix estimation, and homography estimation. The 2D line fitting and 3D plane fitting tasks serve to visualize the behavior of robust estimation across various noise levels. By manually adjusting noise ratios, these tasks elucidates the sampling process inherent in robust methods. For complex scenarios, fundamental and essential matrix estimation tasks assess the applicability of DiffSAC to real-world camera pose estimation. The homography estimation task demonstrates the ability of DiffSAC to handle quasi-convex problems. These tasks provide critical insights into DiffSAC's effectiveness in practical computer vision applications.

\subsection{Experimental Setup}
DiffSAC is implemented based on the DDPM framework \cite{ho2020denoising}. For efficient training with multiple data points, the system constructs batches by concatenating diverse point sets. The diffusion inference process involves $T = 100$ iterative refinement steps. The models undergo training for 100 epochs, employing the Adam optimizer, configured $\beta_1 = 0.9$ and $\beta_2 = 0.999$. The learning rate is initially set to 0.0001 and subsequently diminishes following a cosine annealing schedule. Notably, the noise in the data varies with the actual task, while the noise in the diffusion model to generate confidence is sampled from a standard Gaussian distribution. During inference, each input dataset is replicated $\kappa=20$ times to generate $\kappa$ candidate minimum sets concurrently. These sets are then processed by a sample consensus module to identify the optimal hypothesis, with inlier count as the default consensus metric. The inliers of the best hypothesis are used for final model refinement. To accelerate inference, DiffSAC incorporates DPM-Solver++ \cite{lu2022dpm} as a diffusion model accelerator.

All experiments are conducted on a Linux server equipped with an Intel i7 5.0 GHz CPU and an RTX 4090 GPU. The implementation is developed using PyTorch.

\subsection{2D Line Fitting Task}
\begin{table*}[t]
\centering
\caption{\textbf{2D line fitting.} The mAA@$0.5^{\circ}$ and median error($^{\circ}$) across outlier rates from 10\% to 80\% are reported.}
\setlength{\tabcolsep}{0.1mm}
\renewcommand\arraystretch{1.0}
\resizebox{0.99\textwidth}{!}{
\begin{tabular}{l|cc|cc|cc|cc|cc|cc|cc|cc|c}
\toprule
\multirow{2}{*}{Method} & \multicolumn{2}{c|}{0.1} & \multicolumn{2}{c|}{0.2} & \multicolumn{2}{c|}{0.3} & \multicolumn{2}{c|}{0.4} & \multicolumn{2}{c|}{0.5} & \multicolumn{2}{c|}{0.6} & \multicolumn{2}{c|}{0.7} & \multicolumn{2}{c|}{0.8} & \multirow{2}{*}{Speed (Hz) ↑} \\ \cmidrule{2-17} 
                        & mAA ↑    & Mid. ↓       & mAA ↑       & Mid. ↓    & mAA ↑       & Mid. ↓       & mAA ↑        & Mid. ↓      & mAA ↑       & Mid. ↓       & mAA ↑       & Mid. ↓  & mAA ↑       & Mid. ↓  & mAA ↑       & Mid. ↓  &  \\ \hline  %
Theil-Sen \cite{theil1950rank}    & 0.80 & - & 0.77 & - & 0.75 & - & 0.15 & - & 0.09 & - & 0.06 & - & 0.01 & - & 0.00 & - & 80 \\
RANSAC \cite{fischler1981random}    & 0.86 & 0.05& 0.83 & 0.06 & 0.81 & 0.06 & 0.80 & 0.07 & 0.78 & 0.07 & 0.76 & 0.09 & 0.57 & 0.24 & 0.30 & 0.31 & \textbf{100}\\
DGSAC \cite{tiwari2018dgsac} & 0.86 & 0.05& 0.84 & 0.06 & 0.83 & 0.06 & 0.82 & 0.06 & 0.79 & 0.07 & 0.77 & 0.08 & 0.60 & 0.20 & 0.36 & 0.27 & \textbf{82}\\
CLNet \cite{ji2021clnet}    & 0.87 & 0.05& 0.85 & 0.05 & 0.84 & 0.06 & 0.83 & 0.06 & 0.81 & 0.07 & 0.79 & 0.08 & 0.64 & 0.19 & 0.38 & 0.26 & 60(GPU) 33(CPU)\\
\textbf{Ours (DiffSAC)}     & \textbf{0.89} & \textbf{0.03} & \textbf{0.88} & \textbf{0.04} & \textbf{0.87} & \textbf{0.04} & \textbf{0.87} & \textbf{0.05} & \textbf{0.86} & \textbf{0.05} & \textbf{0.84} & \textbf{0.07} & \textbf{0.70} & \textbf{0.13} & \textbf{0.43} & \textbf{0.20} & 50(GPU) 30(CPU)\\ \bottomrule
\end{tabular}
}
\label{tab:line_result}
\end{table*}

\begin{table}[t]\footnotesize
\centering
\caption{\textbf{2D line fitting on different outlier scales.} The mAA@$0.5^{\circ}$ and median error($^{\circ}$) with 0.5 outlier rates at various outlier scales are reported.}
\setlength{\tabcolsep}{0.1mm}
\renewcommand\arraystretch{1.0}
\begin{tabular}{l|cc|cc|cc}
\toprule
\multirow{2}{*}{Method} & \multicolumn{2}{c|}{0.05} & \multicolumn{2}{c|}{0.1} & \multicolumn{2}{c}{0.2} \\ \cmidrule{2-7} 
                        & mAA ↑    & Mid. ↓       & mAA ↑       & Mid. ↓    & mAA ↑       & Mid. ↓       \\ \hline  %
Theil-Sen \cite{theil1950rank}    & 0.21 & - & 0.09 & - & 0.03 & - \\
RANSAC \cite{fischler1981random}    & 0.80 & 0.06& 0.78 & 0.07 & 0.73 & 0.13 \\
DGSAC \cite{tiwari2018dgsac} & 0.88 & 0.04& 0.79 & 0.07 & 0.74 & 0.12 \\
CLNet \cite{ji2021clnet}    & 0.88 & 0.05& 0.81 & 0.07 & 0.75 & 0.10\\
\textbf{Ours (DiffSAC)}     & \textbf{0.93} & \textbf{0.02} & \textbf{0.86} & \textbf{0.05} & \textbf{0.81} & \textbf{0.08} \\ \bottomrule
\end{tabular}
\label{tab:line_scales}
\end{table}

\begin{table}[t]\footnotesize
\centering
\caption{\textbf{2D line fitting on point-aggregated outliers.} The mAA@$0.5^{\circ}$ and median error($^{\circ}$) with 0.5 outlier rates are reported.}
\setlength{\tabcolsep}{6.0mm}
\renewcommand\arraystretch{1.0}
\begin{tabular}{l|cc}
\toprule

 Method          & mAA ↑    & Mid. ↓   \\ \hline  %
Theil-Sen \cite{theil1950rank}    & 0.03 & - \\
RANSAC \cite{fischler1981random}    & 0.68 & 0.18\\
DGSAC \cite{tiwari2018dgsac} & 0.71 & 0.14\\
CLNet \cite{ji2021clnet}    & 0.72 & 0.14\\
\textbf{Ours (DiffSAC)}     & \textbf{0.80} & \textbf{0.09} \\ \bottomrule
\end{tabular}
\label{tab:line_point}
\end{table}

We first test DiffSAC in 2D line fitting, a basic task in computer vision. We generate synthetic datasets of points in a 2D space, $\chi = \left\{ {\left[ {{x_i},{y_i}} \right]\left| {\ i = 1,2,...,N} \right.} \right\} \in {\mathbb{R}^{N \times 2}}$, where $N$ is the total number of points. For each dataset, we establish a ground truth line within a $10\times10$ picture. Inliers, representing points belonging to the ground truth line, are then positioned along this line. While outliers are randomly distributed across the picture to simulate realistic noise scenarios. The proportion of outliers is systematically varied to evaluate robustness at different contamination levels. Consistent with the definition of a line requiring a minimum of two points, DiffSAC identifies sets of $\gamma =2$ points as minimum sets. We configure DiffSAC with an inlier distance threshold of $\varepsilon = 0.1$, reflecting the perturbation range of inlier points around the ground truth line. In all experiments, we use datasets of $N=100$ points for both training and testing phases.

Quantitative performance is evaluated using the mean Average Accuracy (mAA) metric \cite{barath_learning_2022}. This metric quantifies the angular difference between estimated and ground truth lines. An estimate is accurate if the angular error is within $0.5^\circ$. Table \ref{tab:line_result} summarizes the mAA scores achieved by DiffSAC and comparative methods across varying outlier rates.

As shown in Table \ref{tab:line_result}, DiffSAC consistently achieves superior performance compared to other methods across all tested outlier rates. In scenarios with low outlier contamination, CLNet \cite{ji2021clnet}, RANSAC \cite{fischler1981random}, and Theil-Sen \cite{theil1950rank} exhibit reasonable line estimation accuracy. However, DiffSAC still demonstrates a slight performance advantage with competitive speed. Notably, in high-outlier scenarios, DiffSAC is substantially ahead, maintaining significantly higher mAA scores and demonstrating robust performance. In contrast, the performance of other methods, particularly Theil-Sen, degrades considerably in the presence of increased noise. This indicates that DiffSAC effectively identifies minimum sets that are closely aligned with the ground truth line, rendering it more resilient to high levels of noise. In addition, TABLE \ref{tab:line_scales} quantitatively evaluates the performance of the various methods at different noise scales on 0.5 outlier rate. The results show that DiffSAC exhibits better performance at different noise scales compared to other methods. This experiment demonstrates the robustness of DiffSAC to noise scales. Furthermore, TABLE \ref{tab:line_point} shows the performance of each method under the interference of non-uniform point-aggregated noise. Due to the effective extraction of data semantics by DiffSAC's neural network, DiffSAC still achieves better performance in this interference situation.

\begin{figure*}[t]
  \centering
   \includegraphics[width=0.80\linewidth]{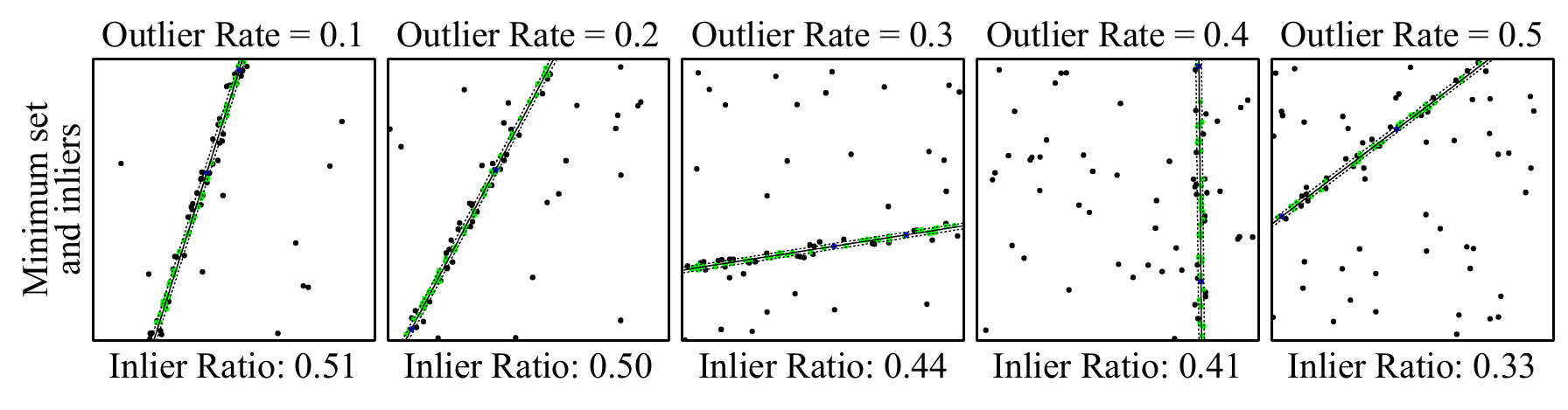}
    \vspace{-0.1cm}
   \caption{\textbf{The qualitative results of DiffSAC on 2D line fitting.} \textbf{\textcolor{green!50!gray}{green}}, \textbf{black} respectively indicate \textbf{\textcolor{green!50!gray} {inliers}}, and \textbf{outliers}. DiffSAC can fit the accurate 2D line at various outlier rates.}
   \label{fig:line}
\end{figure*}

\begin{figure*}[t]
  \centering
   \includegraphics[width=0.80\linewidth]{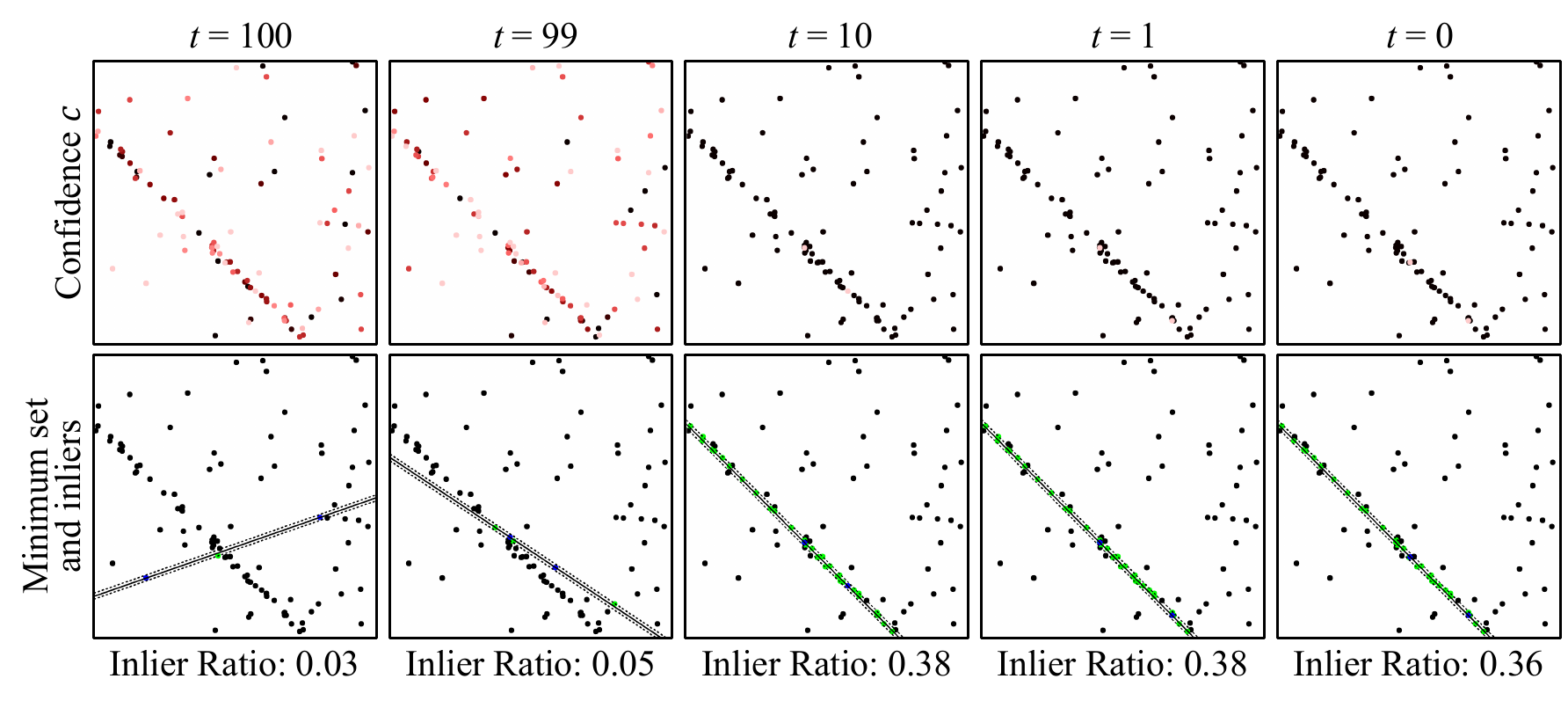}
    \vspace{-0.1cm}
   \caption{\textbf{The step results of DiffSAC on 2D line fitting.} The brighter the points in confidence $c$, the higher their confidence value. The other colors are set the same as Fig. \ref{fig:line}. DiffSAC can iteratively refine the bad initial confidence $c$ sampled from noise to a high-quality minimum set at the 0.5 outlier rate.}
   \label{fig:line_step}
\end{figure*}

Complementing the quantitative analysis, Fig. \ref{fig:line} presents qualitative results, visually illustrating the lines estimated by DiffSAC under varying noise levels. These visual results align with the quantitative findings, confirming the ability of DiffSAC to accurately estimate lines even when substantial noise is present. Furthermore, Fig. \ref{fig:line_step} visualizes the iterative refinement process within DiffSAC, specifically in a scenario with a 0.5 outlier rate. The visualization tracks the evolution of confidence $c$, demonstrating that DiffSAC can effectively refine an initially noisy set of points to converge on a minimum set with high confidence. This rapid convergence highlights the capacity of DiffSAC to leverage the generative power of diffusion models for identifying high-quality hypotheses. The progressive refinement also showcases the ability of DiffSAC to perform local optimization, further enhancing the quality of the identified minimum set once a promising candidate is found. These qualitative observations emphasize the effectiveness of DiffSAC in exploiting geometric properties to guide the generation of accurate confidence $c$ for minimum sets, especially in challenging high-noise conditions.

\subsection{3D Plane Fitting Task}
\label{plane}
We further assess the effectiveness of DiffSAC on 3D plane fitting, a task with broader applicability.  The experimental setup mirrors the 2D line fitting task, but now utilizes three-dimensional data points, $\chi = \left\{ {\left[ {{x_i},{y_i},{z_i}} \right]\left| {\ i = 1,2,...,N} \right.} \right\} \in {\mathbb{R}^{N \times 3}}$. Consistent with the geometric definition of a plane, DiffSAC selects minimum sets of $\gamma =3$ points. We again use synthetic datasets with $N=100$ points for both training and testing, unless otherwise specified.

\begin{figure}[t]
  \centering
   \includegraphics[width=0.99\linewidth]{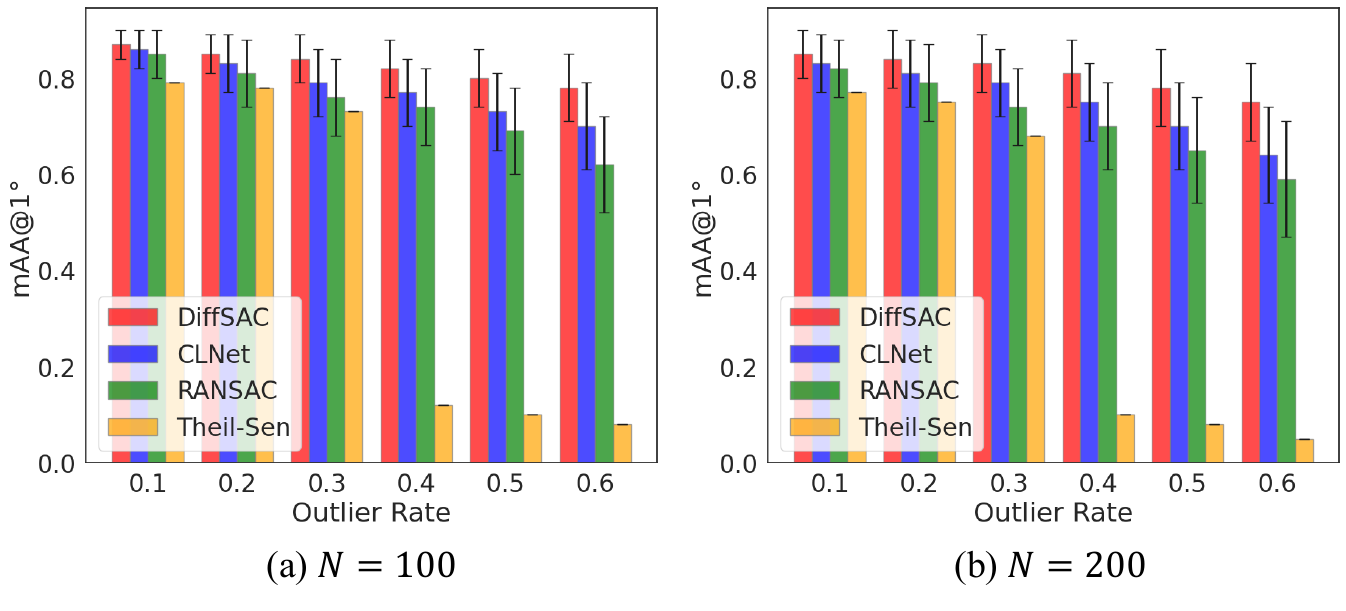}
    \vspace{-0.5cm}
   \caption{\textbf{3D Plane fitting on (a) 100 and (b) 200 synthetic data points with various outlier rates.} The mAA@$1^{\circ}$ is reported, and higher is better. DiffSAC achieves the best performance at outlier rates from 10\% to 60\%.}
   \label{fig:plane_bar}
\end{figure}

\begin{figure*}[t]
  \centering
   \includegraphics[width=0.80\linewidth]{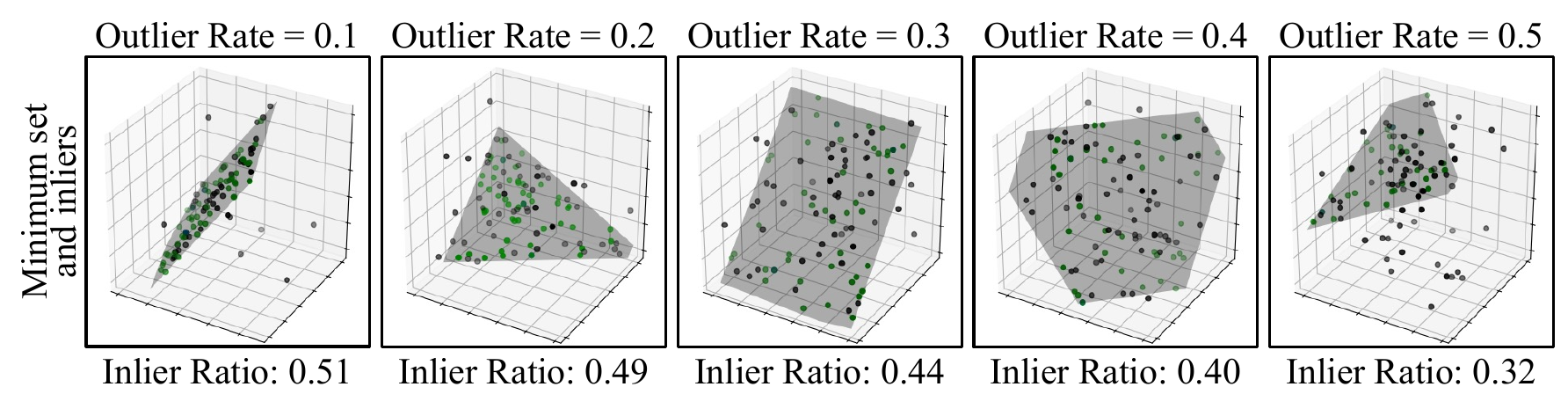}
    \vspace{-0.1cm}
   \caption{\textbf{The qualitative results of DiffSAC on 3D Plane fitting.} \textbf{\textcolor{green!50!gray}{green}}, \textbf{black} respectively indicate \textbf{\textcolor{green!50!gray} {inliers}}, and \textbf{outliers}. DiffSAC can fit the accurate 3D plane at various outlier rates.}
   \label{fig:plane}
\end{figure*}

Quantitative evaluation of 3D plane fitting performance utilizes the mAA metric, adjusted for angular error between planes with $1^\circ$ tolerance. Fig. \ref{fig:plane_bar} (a) presents the mAA results, demonstrating that DiffSAC outperforms other methods across all outlier rates. To further examine the robustness and generalization capabilities of DiffSAC, we evaluate its performance on datasets with doubled points (N=200), without any additional training. Fig. \ref{fig:plane_bar} (b) showcases the strong generalization of DiffSAC, maintaining high performance with increased data point density. Qualitative visualizations of the fitted planes, presented in Fig. \ref{fig:plane}, further corroborate these quantitative findings. These visualizations demonstrate that DiffSAC can accurately fit planes even under high levels of noise.

\subsection{Fundamental Matrix Estimation Task}
\label{Fundamental}
Fundamental matrix estimation is a crucial component in computer vision, which inputs correspondences between pairs of images and can be decomposed to obtain the camera pose. In this study, we use the dataset and experimental setup from the CVPR 2020 RANSAC tutorial \cite{barath2020ransac}. Its training set comprises 12 scenes and over one million image pairs. The test set consists of 2 scenes with 4,950 image pairs per scene. These correspondences are extracted using RootSIFT \cite{arandjelovic_three_2012} correspondences, which are matched using nearest neighbor search. For each correspondence, a 128-dimensional SIFT descriptor is extracted. The geometric feature is therefore defined as $\chi  = \left\{ {\left[ {x_1^i,y_1^i,des_1^i,x_2^i,y_2^i,des_2^i} \right]\left| {i = 1,2,...,N} \right.} \right\} \in {\mathbb{R}^{N \times 260}}$. Hypotheses are solved using the 8-point algorithm \cite{longuet1981computer}, with an inlier threshold of $\varepsilon=4$.

The quantitative performance of our method is summarized in Table \ref{tab:f_result}. We use the mAA metric to evaluate the camera pose accuracy, specifically focusing on rotation and translation. Our method achieves improved results, exhibiting lower errors compared to other approaches. This enhanced performance can be attributed to the integration of a diffusion model within our estimation process. This integration allows for a more effective evaluation of potential solutions, ultimately leading to the selection of a superior fundamental matrix.

\begin{figure}[t]
  \centering
   \includegraphics[width=0.99\linewidth]{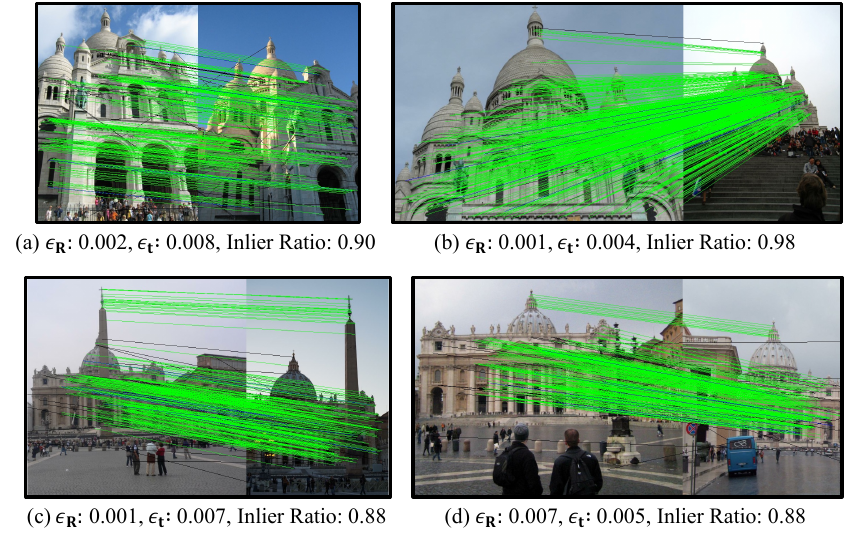}
    \vspace{-0.5cm}
   \caption{\textbf{The qualitative results of DiffSAC on the fundamental matrix estimation task.} The rotation, translation errors, and inlier rate are reported. \textbf{\textcolor{green!50!gray}{green}}, \textbf{black} respectively indicate \textbf{\textcolor{green!50!gray} {inliers}}, and \textbf{outliers}.} 
   \label{fig:f}
\end{figure}

\begin{figure}[t]
  \centering
   \includegraphics[width=0.99\linewidth]{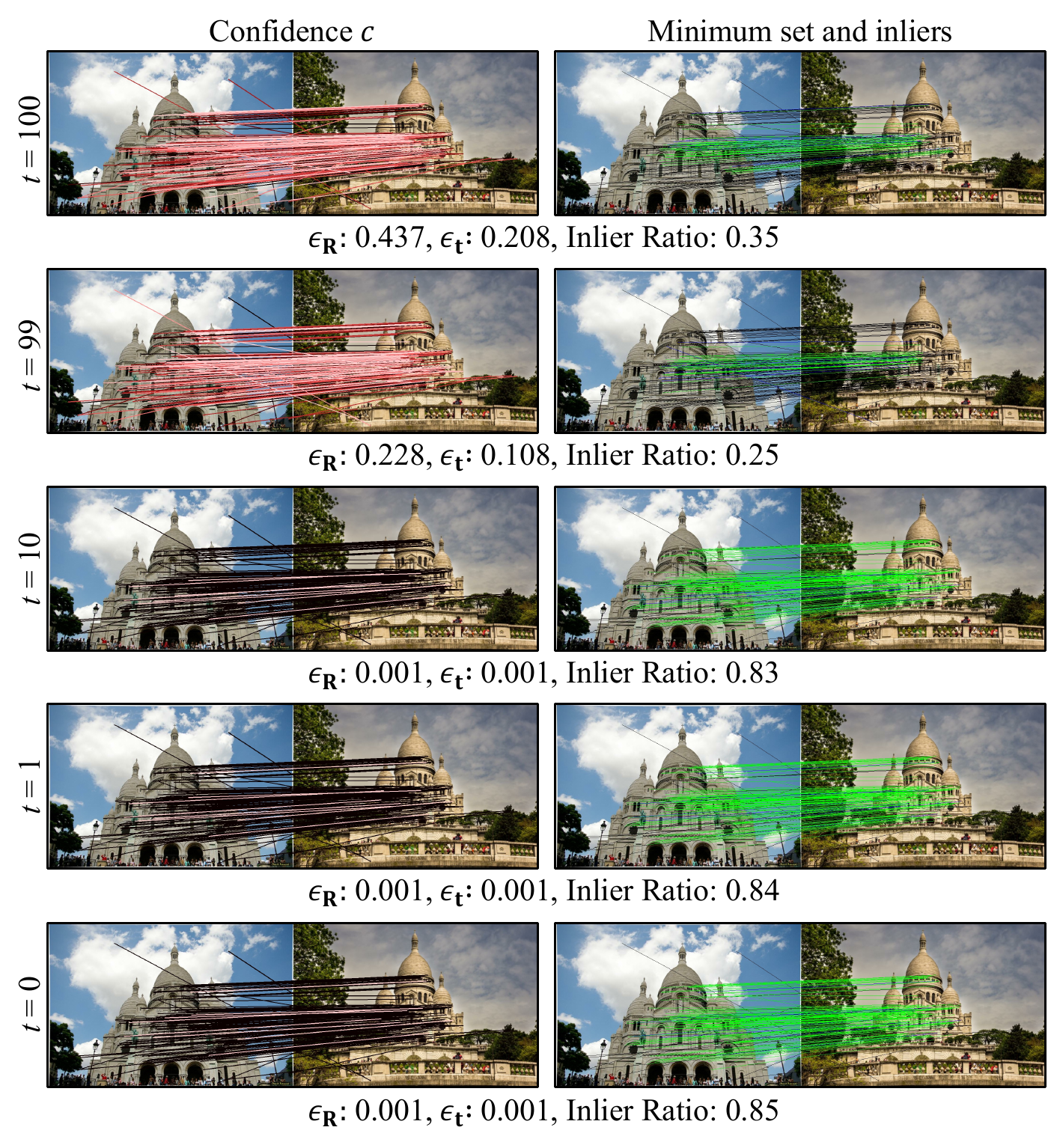}
    \vspace{-0.1cm}
   \caption{\textbf{The step results of DiffSAC on the fundamental matrix estimation task.} The brighter the lines in confidence $c$, the higher confidence value. The other colors of the lines and the evaluation metrics are consistent with Fig. \ref{fig:f}.} 
   \label{fig:f_step}
\end{figure}

Qualitative results, visually demonstrating the effectiveness of our method, are presented in Fig. \ref{fig:f}. These visualizations reveal that our approach effectively disregards inaccurate and structurally irrelevant correspondences, leading to a more precise fundamental matrix estimation. Furthermore, Fig. \ref{fig:f_step} illustrates the step-by-step refinement of the estimation process. This figure demonstrates a gradual reduction in both rotation and translation errors as the process progresses. This visual evidence suggests that our method iteratively improves the solution quality by generating and selecting high-quality sets of correspondences while discarding bad sets.

\begin{table*}[t]
\centering
\caption{\textbf{Fundamental matrix estimation.} The mAA@$10^{\circ}$ and median error of rotation and direction of translation are reported.}
\setlength{\tabcolsep}{4.5mm}
\renewcommand\arraystretch{1.0}
\footnotesize
\begin{tabular}{l|cc|cc|c}
\toprule
\multirow{2}{*}{Method} & \multicolumn{2}{c|}{mAA@$10^{\circ}$ ↑} & \multicolumn{2}{c|}{Median ($^{\circ}$) ↓} & \multirow{2}{*}{Speed (Hz) ↑} \\ \cmidrule{2-5} 
                        & \textbf{R}        & \textbf{t}        & ${\epsilon_\textbf{R}}$        & ${\epsilon_\textbf{t}}$  & \\ \hline  %
RANSAC \cite{fischler1981random}            & 0.681            & 0.371            & 2.952        & 8.349  & 31 \\
LO-RANSAC \cite{chum2003locally}            & 0.690            & 0.403            & 2.579        & 7.625 & 27 \\
USAC \cite{raguram_usac_2013}               & 0.698            & 0.439            & 2.014        & 6.415 & 28  \\
DGSAC \cite{tiwari2018dgsac}                & 0.713         & 0.504             & 1.875        & 4.947  & 24 \\
NG-RANSAC \cite{brachmann2019neural}        & 0.718         & 0.574            & 1.587        & 2.918 & 22(GPU) 8(CPU)  \\
MAGSAC++ \cite{barath_marginalizing_2022}     & 0.723    & 0.585            & 1.476        & 2.632  & \textbf{53} \\

\textbf{Ours (DiffSAC)}                    & \textbf{0.783}             & \textbf{0.641}            & \textbf{0.886}            & \textbf{1.819}  & 30(GPU) 13(CPU) \\ \bottomrule
\end{tabular}
\label{tab:f_result}
\end{table*}

\begin{figure}[t]
  \centering
   \includegraphics[width=0.99\linewidth]{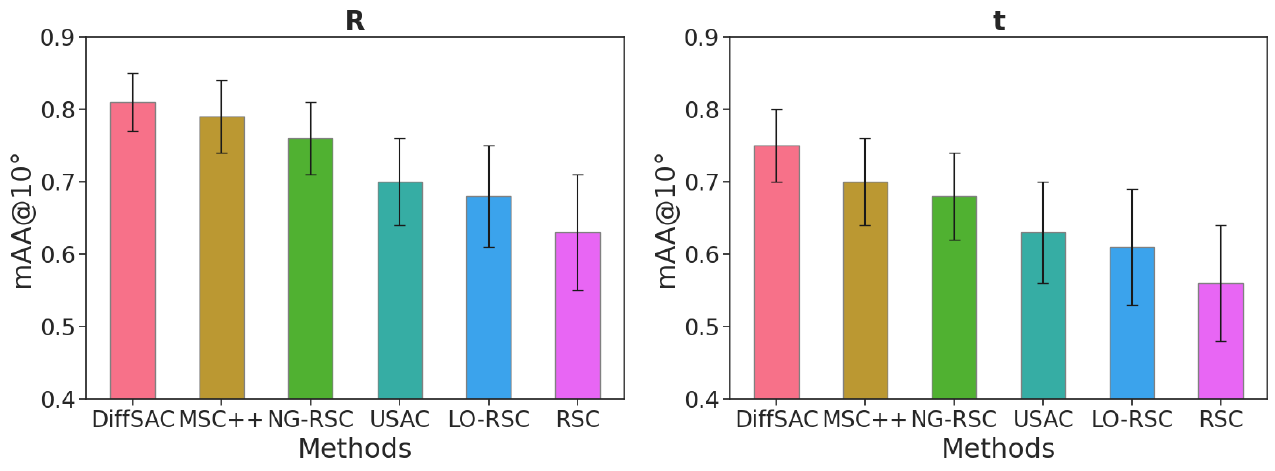}
    \vspace{-0.4cm}
   \caption{\textbf{Fundamental matrix estimation on the KITTI dataset.} The mAA@$10^{\circ}$ of of rotation and direction of translation are reported.}
   \label{fig:kitti_bar}
\end{figure}

\begin{figure*}[t]
  \centering
   \includegraphics[width=0.99\linewidth]{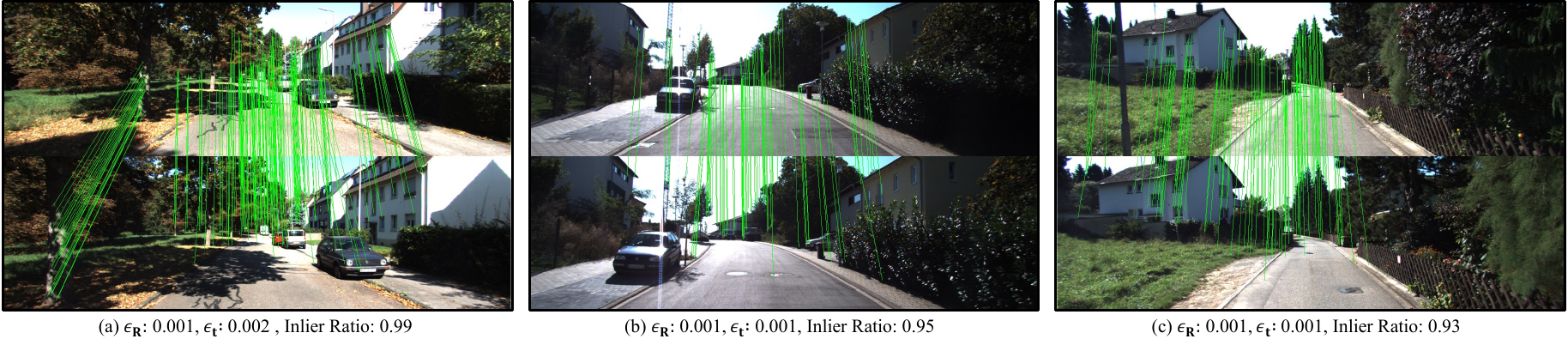}
    \vspace{-0.1cm}
   \caption{\textbf{The DiffSAC qualitative results of fundamental matrix estimation on the KITTI dataset.} The inliers are indicated by the green line. DiffSAC can estimate the accurate fundamental matrix on the KITTI dataset.}
   \label{fig:kitti}
\end{figure*}

To further validate our findings and assess the robustness of DiffSAC across different scenarios, we conduct experiments on the KITTI odometry dataset. Following standard practice, we use sequences 00-07 for training and sequences 08-10 for testing. Feature points are computed and matched using the SIFT algorithm. The quantitative outcomes, depicted in Fig. \ref{fig:kitti_bar}, show that our method achieves superior performance in both rotation and translation accuracy metrics on this dataset. These results reinforce our earlier conclusions about the effectiveness of our approach and highlight its adaptability to diverse situations. Qualitative results presented in Fig. \ref{fig:kitti} demonstrate that our method successfully estimates accurate fundamental matrices on the KITTI dataset, enabling the derivation of reliable camera poses. These experiments demonstrate the robustness of DiffSAC to correspondences from different scenarios. This can be attributed to the diffusion model in DiffSAC that can effectively utilize the pixel location and descriptor information of correspondences.

To provide a more granular comparison of fundamental matrix estimation algorithms, particularly in scenarios with varying levels of data corruption, we conduct a controlled experiment using a synthetic dataset. This dataset is generated from the ModelNet40 dataset. By virtually positioning a camera and rendering a chosen CAD model from different viewpoints, we synthesize pairs of images. Each image pair in this synthetic dataset is designed to contain $N=200$ feature point correspondences, with random noise to simulate varying outlier contamination rates. The results of this experiment, detailed in Table \ref{tab:ModelNet40}, demonstrate a consistent trend. Our method, DiffSAC, consistently achieves the most accurate rotation and translation estimation as the proportion of outlier correspondences increases. In simplified situations characterized by low outlier rates, all methods exhibit generally strong performance. However, as outlier rates become more substantial, the performance of other methods degrades noticeably, while DiffSAC maintains a robust level of accuracy. This sustained performance is attributed to the ability of DiffSAC to consistently identify high-quality minimum sets of correspondences, enabling the robust estimation of the fundamental matrix even in the presence of significant data corruption. This experiment further underscores the pronounced robustness of DiffSAC when confronted with varying degrees of outlier contamination.

\begin{table*}[t]
\centering
\caption{\textbf{Fundamental matrix estimation on ModelNet40 dataset.} The mAA@$10^{\circ}$ ↑ of rotation and direction of translation at outlier rates from 10\% to 60\% are reported.}
\setlength{\tabcolsep}{3.0mm}
\renewcommand\arraystretch{0.9}
\resizebox{0.99\textwidth}{!}{
\begin{tabular}{l|cc|cc|cc|cc|cc|cc}
\toprule
\multirow{2}{*}{Method} & \multicolumn{2}{c|}{0.1} & \multicolumn{2}{c|}{0.2} & \multicolumn{2}{c|}{0.3} & \multicolumn{2}{c|}{0.4} & \multicolumn{2}{c|}{0.5} & \multicolumn{2}{c}{0.6}  \\ \cmidrule{2-13} 
    & \textbf{R}    & \textbf{t}    & \textbf{R}    & \textbf{t}  & \textbf{R}    & \textbf{t}    & \textbf{R}    & \textbf{t}  & \textbf{R}    & \textbf{t} & \textbf{R}    & \textbf{t} \\ \hline  %
RANSAC \cite{fischler1981random}          & 0.917 & 0.753 & 0.849 & 0.726 & 0.778 & 0.668 & 0.699 & 0.566 & 0.602 & 0.466 & 0.418 & 0.376 \\
LO-RANSAC \cite{chum2003locally}          & 0.928 & 0.764 & 0.861 & 0.743 & 0.793 & 0.681 & 0.716 & 0.579 & 0.629 & 0.484 & 0.453 & 0.399\\
USAC \cite{raguram_usac_2013}             & 0.937 & 0.776 & 0.872 & 0.759 & 0.804 & 0.694 & 0.728 & 0.593 & 0.643 & 0.502 & 0.487 & 0.414  \\
DGSAC \cite{tiwari2018dgsac} & 0.940 & 0.781 & 0.879 & 0.767 & 0.809 & 0.700 & 0.733 & 0.601 & 0.651 & 0.508 & 0.499 & 0.420  \\
NG-RANSAC \cite{brachmann2019neural}      & 0.942 & 0.785 & 0.880 & 0.772 & 0.812 & 0.702 & 0.742 & 0.605 & 0.658 & 0.514 & 0.508 & 0.427  \\
MAGSAC++ \cite{barath_marginalizing_2022} & 0.950 & 0.799 & 0.891 & 0.781 & 0.820 & 0.712 & 0.750 & 0.615 & 0.670 & 0.522 & 0.524 & 0.438 \\
\textbf{Ours (DiffSAC)}     & \textbf{0.965} & \textbf{0.813} & \textbf{0.909} & \textbf{0.790} & \textbf{0.835} & \textbf{0.724} & \textbf{0.763} & \textbf{0.628} & \textbf{0.684} & \textbf{0.537} & \textbf{0.547} & \textbf{0.453} \\ \bottomrule
\end{tabular}
}
\label{tab:ModelNet40}
\end{table*}

\subsection{Essential Matrix Estimation Task}
Essential matrix estimation is crucial for determining camera pose, utilizing intrinsic camera parameters and epipolar geometry to compute the relative rotation and translation between two images in normalized coordinates. Following the experimental protocol established for fundamental matrix estimation, we train and evaluate DiffSAC on the same dataset and setup. The input to DiffSAC consists of geometric features, defined as $\chi  = \left\{ {\left[ {x_1^i,y_1^i,des_1^i,x_2^i,y_2^i,des_2^i} \right]\left| {i = 1,2,...,N} \right.} \right\} \in {\mathbb{R}^{N \times 260}}$. These features encapsulate the coordinates and descriptors of matched feature points between image pairs. For hypothesis generation, DiffSAC employs the efficient 5-point algorithm \cite{nister2004efficient}, a standard technique in essential matrix estimation. We set the inlier threshold to $\varepsilon=1$ to distinguish between inlier and outlier correspondences.

\begin{table*}[t]\footnotesize 
\centering
\caption{\textbf{Essential matrix estimation.} The mAA@$10^{\circ}$ and median error of rotation and direction of translation are reported.}
\setlength{\tabcolsep}{4.5mm}
\renewcommand\arraystretch{1.0}
\begin{tabular}{l|cc|cc|c}
\toprule
\multirow{2}{*}{Method} & \multicolumn{2}{c|}{mAA@$10^{\circ}$ ↑} & \multicolumn{2}{c|}{Median ($^{\circ}$) ↓} & \multirow{2}{*}{Speed (Hz) ↑} \\ \cmidrule{2-5}
                        & \textbf{R}        & \textbf{t}        & ${\epsilon_\textbf{R}}$        & ${\epsilon_\textbf{t}}$  & \\ \hline  %
RANSAC \cite{fischler1981random}            & 0.701            & 0.415            & 1.651        & 6.374 & 50 \\
LO-RANSAC \cite{chum2003locally}            & 0.716            & 0.422            & 1.421        & 6.106 & 46 \\
USAC \cite{raguram_usac_2013}               & 0.723            & 0.426            & 1.396        & 5.932   & 48 \\
NG-RANSAC \cite{brachmann2019neural}     & 0.753    & 0.530            & 1.287        & 2.792   & 38(GPU) 17(CPU)\\
MAGSAC++ \cite{barath_marginalizing_2022}     & 0.778    & 0.553            & 1.195        & 2.284   & \textbf{86} \\

\textbf{Ours (DiffSAC)}                    & \textbf{0.798}             & \textbf{0.651}            & \textbf{0.863}     & \textbf{1.779} & 48(GPU) 22(CPU) \\ \bottomrule
\end{tabular}
\label{tab:e_result}
\end{table*}

\begin{figure}[t]
  \centering
   \includegraphics[width=0.99\linewidth]{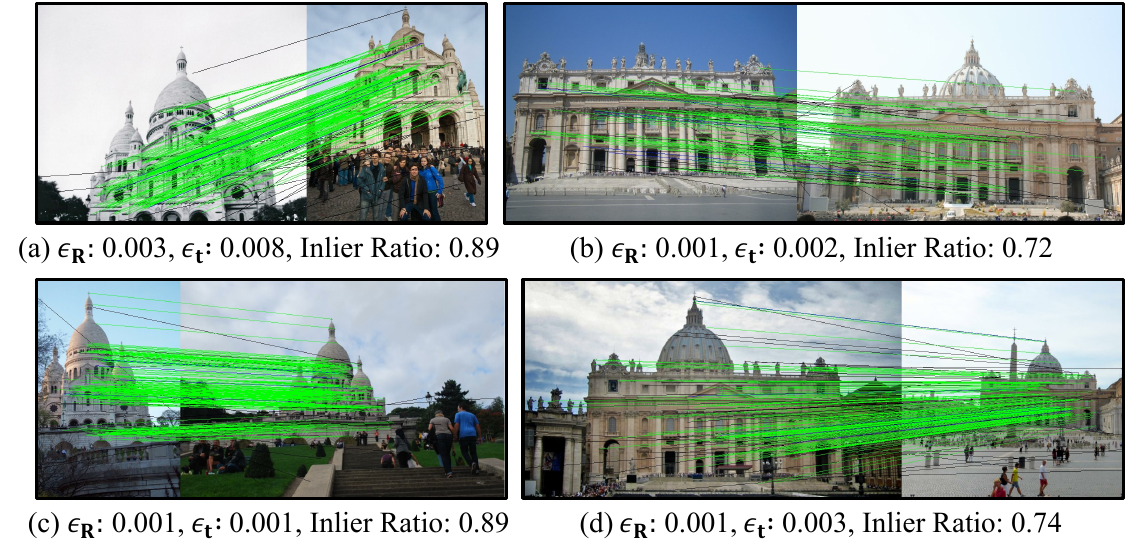}
    \vspace{-0.4cm}
   \caption{\textbf{The qualitative results of DiffSAC on essential matrix estimation task.} The rotation, translation errors, and inlier rate are reported. The colors of lines are set the same as in Fig. \ref{fig:f}.} 
   \label{fig:e}
\end{figure}

\begin{figure}[t]
  \centering
   \includegraphics[width=0.99\linewidth]{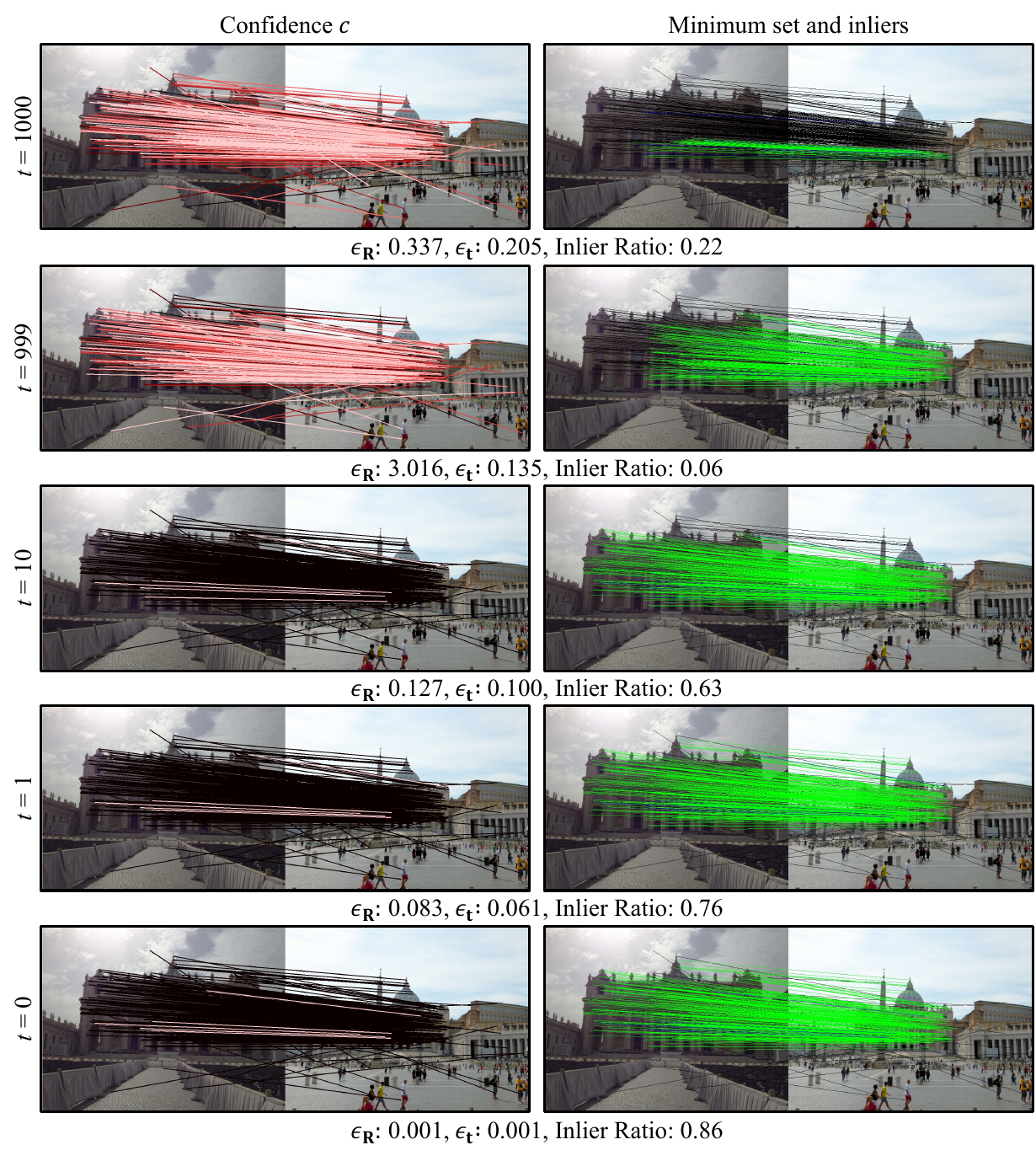}
    \vspace{-0.4cm}	
   \caption{\textbf{The step results of DiffSAC on the essential matrix estimation task.} The colors of lines and evaluation metrics are consistent with Fig. \ref{fig:e}.}
   \label{fig:e_step}
\end{figure}

Quantitative results are summarized in Table \ref{tab:e_result}. The data reveals that DiffSAC achieves superior performance compared to other robust estimation methods in terms of accuracy. Qualitative results, visualized in Fig. \ref{fig:e}, further illustrate the capability of DiffSAC to estimate accurate essential matrices across diverse scenes and viewing angles. Moreover, Fig. \ref{fig:e_step}, displaying multi-step essential matrix estimation, demonstrates the ability of DiffSAC to consistently generate high-quality minimum sets of correspondences, crucial for robust estimation. These findings collectively highlight that DiffSAC can estimate accurate essential matrices, effectively leveraging the generative power of diffusion models to produce deterministic high-quality minimum sets for this task.

\subsection{Homography Estimation Task}
\begin{figure}[t]
  \centering
   \includegraphics[width=0.99\linewidth]{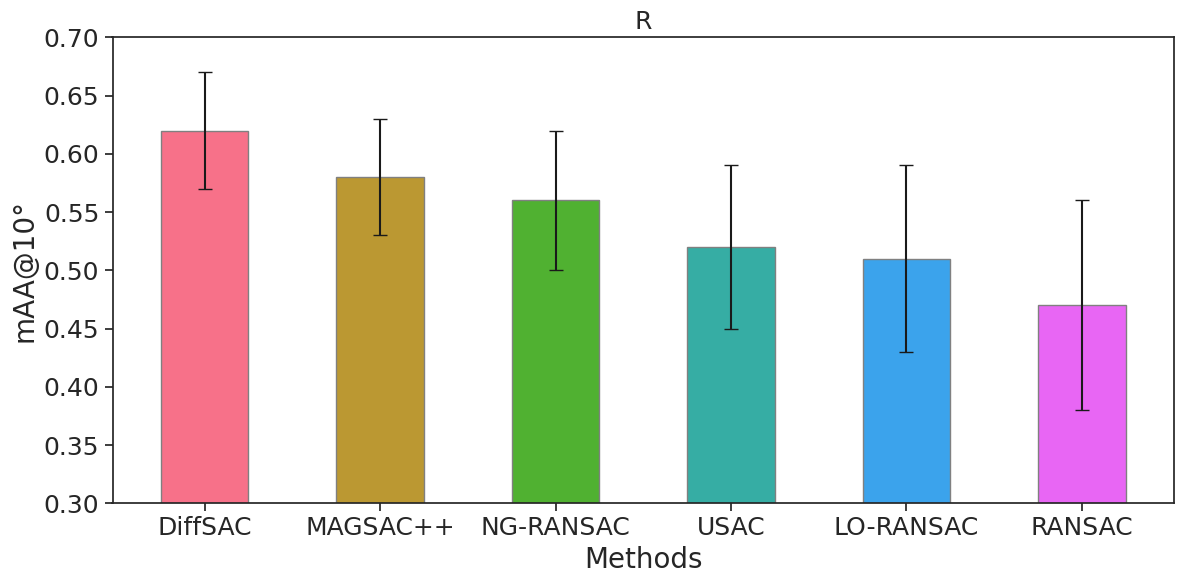}
    \vspace{-0.4cm}
   \caption{\textbf{Homography estimation on the KITTI dataset.} The mAA@$10^{\circ}$ of rotation is reported.}
   \label{fig:kitti_h}
\end{figure}

We further evaluate DiffSAC on the homography estimation task. This task serves to assess the performance of DiffSAC in handling quasi-convex residuals, which are a characteristic of many robust estimation problems. We utilize the KITTI dataset for training and evaluation, maintaining consistency with the experimental settings used in the fundamental matrix estimation experiments detailed in Section \ref{Fundamental}. The geometric feature input $\chi  = \left\{ {\left[ {x_1^i,y_1^i,des_1^i,x_2^i,y_2^i,des_2^i} \right]\left| {i = 1,2,...,N} \right.} \right\} \in {\mathbb{R}^{N \times 260}}$ remains the same. To estimate the homography matrix $H$, DiffSAC utilizes the Direct Linear Transform (DLT) algorithm \cite{hartley1997triangulation}, a widely adopted method for homography estimation. The inlier threshold for this task is set to $\varepsilon=0.1$. The residual, which quantifies the reprojection error of matched points under the estimated homography, is computed as:
\begin{equation}
r=\frac{\left \| (H_{row=1:2} - [x_2,y_2]^TH_{row=3})[x_1,y_1,1]^T \right \| }{H_{row=3}\cdot [x_1,y_1,1]^T}.
\end{equation}
The quantitative results, presented in Figure \ref{fig:kitti_h}, demonstrate that DiffSAC outperforms other methods in homography estimation on the KITTI dataset. This outcome underscores the broader applicability and effectiveness of DiffSAC as a robust estimation technique. This experiment confirms the efficacy of DiffSAC extends beyond specific problem instances like essential and fundamental matrix estimation, showcasing its potential for more general robust estimation challenges.

\subsection{Ablation Study}
To evaluate the contribution of each component within our DiffSAC approach, we performed ablation studies. In these experiments, we systematically altered or removed specific modules of our method, while maintaining consistent experimental data and settings. This allowed us to isolate and understand impacts of elements on the overall performance.

\begin{table*}[t]\scriptsize 
\centering
\caption{\textbf{The ablation study results} of DiffSAC on fundamental matrix estimation.}
\setlength{\tabcolsep}{4.0mm}
\renewcommand\arraystretch{1.1}
\begin{tabular}{c|l|cc|cc}
\toprule
\multirow{2}{*}{Exp.}    & \multirow{2}{*}{Method}                           & \multicolumn{2}{c|}{mAA@$10^{\circ}$ ↑} & \multicolumn{2}{c}{Median ($^{\circ}$) ↓}  \\ \cmidrule{3-6} 
                     &                                                   & \textbf{R}        & \textbf{t}        & ${\epsilon_\textbf{R}}$        & ${\epsilon_\textbf{t}}$    \\\hline %
\multirow{2}{*}{(a)} & Ours (w/ direct learning confidence)                            & 0.687        & 0.393        & 2.761        & 7.835   \\
                     & Ours (full, w/ diffusion model) & \textbf{0.783}             & \textbf{0.641}            & \textbf{0.886}            & \textbf{1.819}       \\ \hline
\multirow{2}{*}{(b)} & Ours (w/ LO-RANSAC sample consensus)  &\textbf{0.794}             & \textbf{0.657}            & \textbf{0.874}            & \textbf{1.763}   \\
                     & Ours (full, w/ RANSAC sample consensus) & 0.783      & 0.641     & 0.886   & 1.819   \\ \hline
\multirow{5}{*}{(c)} & Ours (w/o descriptors)                            & 0.725        & 0.603        & 1.271        & 2.193   \\
                    & Ours (w/ ORB descriptors)                            & 0.751        & 0.623        & 0.997        & 2.002   \\
                    & Ours (w/ SuperPoint descriptors)                            & \textbf{0.787}        & \textbf{0.646}        & \textbf{0.883}        & \textbf{1.813}   \\
                     & Ours (full, w/ SIFT descriptors + Gaussian) & 0.738             & 0.615            & 1.147           & 2.026       \\
                     & Ours (full, w/ SIFT descriptors) & 0.783             & 0.641            & 0.886           & 1.819       \\ \hline
\multirow{2}{*}{(d)} & Ours (w/o probabilistic sampling) & 0.761        & 0.627        & 0.989        & 1.958   \\
                     & Ours (full, w/ max sampling) & \textbf{0.783}             & \textbf{0.641}            & \textbf{0.886}            & \textbf{1.819}       \\ \hline
\multirow{3}{*}{(e)} & Ours (w/ MLPs)  & 0.713             & 0.569            & 1.613            & 2.987   \\
                    & Ours (w/ DGCNN \cite{wang2019dynamic})  & 0.726             & 0.592            & 1.325            & 2.074   \\
                     & Ours (full, w/ Point-e \cite{song2020denoising}) & \textbf{0.783}             & \textbf{0.641}            & \textbf{0.886}            & \textbf{1.819}  \\ \hline
\multirow{4}{*}{(f)} & RANSAC (10k iterations)   & 0.681            & 0.371            & 2.952        & 8.349   \\
& NG-RANSAC (10k iterations)  & 0.718 & 0.574 & 1.587 & 2.918  \\
& MAGSAC++ (10k iterations)   & 0.723 & 0.585 & 1.476 & 2.632   \\
                     & Ours \textbf{(2k iterations)} & \textbf{0.783}             & \textbf{0.641}            & \textbf{0.886}            & \textbf{1.819}       \\ \bottomrule
\end{tabular}
\label{tab:ablation}
\end{table*}

\textbf{The Effect of the diffusion model} is tested in Table \ref{tab:ablation} (a). We attempt to directly train a neural network to predict confidence $c$ from the input data points $\chi$. However, this direct approach yields significantly lower performance and struggles to identify high-quality minimum sets effectively. In contrast, DiffSAC leverages the iterative refinement process of diffusion models to generate confidence $c$ that closely approximates the true confidence. This continuous refinement of the minimum set is crucial for achieving robust performance.

\textbf{The plug-and-play capability of DiffSAC} is demonstrated in Table \ref{tab:ablation} (b). We combined the minimum set estimation of DiffSAC with the local hypothesis refinement of LO-RANSAC \cite{chum2003locally}. By substituting DiffSAC for the standard RANSAC within this established framework, we observed a further performance increase. This improvement highlights the plug-and-play capability of DiffSAC and its compatibility with existing sampling consensus methods. This versatility stems from DiffSAC's unique combination of deep learning and the principles of classical sample consensus.

\textbf{The effect of descriptors} is tested in Table \ref{tab:ablation} (c). Our findings indicate that DiffSAC can accept various descriptors as input. These descriptors can provide rich image information to DiffSAC, offering the diffusion model with crucial contextual understanding of the image content underlying the data points. Without descriptors, relying solely on coordinate information, DiffSAC struggles to converge effectively. In addition, the SuperPoint descriptors are slightly better than the other methods, possibly because the learning-based SuperPoint output descriptors may be better suited for DiffSAC to decode features. Notably, DiffSAC does not specify the type of descriptors. We use the SIFT descriptors that from the public dataset for a fairer comparison of performance. If there is Gaussian noise in the descriptor of SIFT, the performance degradation will be drastically reduced. These results underscore the role of descriptors in guiding the diffusion process towards meaningful solutions.

\textbf{Different sampling approaches} are compared in Table \ref{tab:ablation} (d). Our experiments show that maximum sampling of the minimum set performs marginally better than probabilistic sampling. This subtle advantage arises because the diffusion model generates highly accurate and deterministic confidence $c$, as visualized in Figure \ref{fig:f_step}. Probabilistic sampling, in contrast, introduces the possibility of selecting less optimal minimum sets, potentially hindering performance.

\textbf{Different neural networks} are tested in Table \ref{tab:ablation} (e). Compared to MLPs and DGCNN-based networks, the transformer-based network demonstrates superior feature extraction from the data points. The ``attention" mechanism inherent in transformers proves particularly effective in discerning the complex relationships between confidence $c$ and data points $\chi$, ultimately leading to enhanced performance.

\begin{figure}[t]
  \centering
   \includegraphics[width=0.99\linewidth]{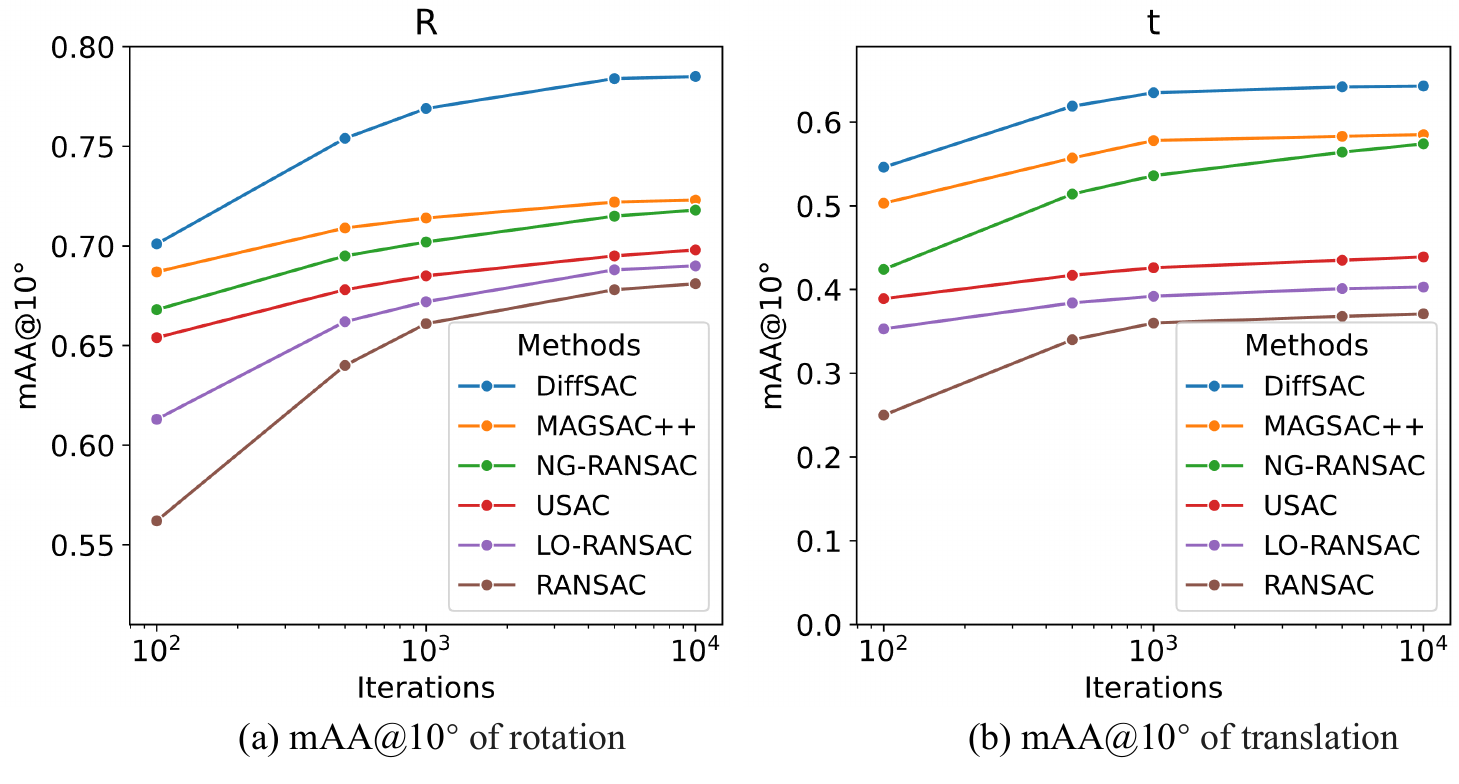}
    \vspace{-0.4cm}
   \caption{\textbf{The mAA@$10^{\circ}$ ↑ of various methods at different iterations.} The rotation and translation errors are reported. DiffSAC achieves excellent performance with few iterations, which proves its high efficiency.}
   \label{fig:iteration}
\end{figure}

\textbf{Sampling efficiency} is compared in Table \ref{tab:ablation} (f) and Fig. \ref{fig:iteration}. By employing DPM-Solver++ \cite{lu2022dpm} to accelerate the inference of diffusion models, DiffSAC achieves rapid hypothesis generation under GPU parallel computing. It solves for high-quality hypotheses in just 33 milliseconds with 2000 iterations, encompassing both diffusion sampling and evaluate consensus. The data preprocessing and feature embedding take 12\% of the time, and diffusion sampling takes 75\% of the time. Benefiting from the small number of high-quality hypotheses to be evaluated, evaluate consensus only takes 13\% of the time. The GPU memory usage is around 2GB for inference. Moreover, our results with varying iteration counts (Fig. \ref{fig:iteration}) reveal that DiffSAC consistently achieves superior performance with significantly fewer iterations than competing methods. For instance, with only 2,000 iterations, DiffSAC surpasses the accuracy of methods like MAGSAC++ that use over 10,000 iterations. This efficiency gain stems from the diffusion model's ability to preemptively filter out bad minimum sets before hypothesis solving, focusing computational resources only on promising candidates. By effectively trading numerous random samples for a small batch of guided, high-quality ones, DiffSAC provides a more efficient and robust estimation pipeline.

\section{Discussion}
\label{discussion}
The learning-based robust estimation methods directly in a one-shot manner \cite{poursaeed2018deep} often function as black boxes. This characteristic obscures the process by which these methods arrive at their results, hindering interpretability. This limited interpretability raises concerns about the practicality of such approaches in engineering applications, where understanding the behavior of methods is often crucial. In contrast, traditional methods grounded in mathematical theory, such as optimization-based model fitting, offer clear interpretability and well-defined application scopes. However, many of these traditional methods struggle to effectively incorporate geometric data features and can be inefficient in data sampling.

Recognizing these complementary strengths, integrating traditional and learning-based methods emerges as a promising direction to achieve both interpretability and high performance in robust estimation. DiffSAC embodies this hybrid approach. By refining confidence $c$ based on geometric features within a diffusion model, DiffSAC outputs high-quality minimum sets. This framework effectively merges the advantages of deep learning with the established principles of sample consensus processes. This integration enhances the interpretability of DiffSAC by providing a clearer understanding of how geometric information contributes to the final estimation.

Furthermore, DiffSAC can easily be applied to other robust estimation tasks due to the following reasons:
\begin{itemize}
	\item DiffSAC takes data points as input to predict a vector representing the confidence of each data point in the minimum set. This implies DiffSAC can process various types of information beyond just coordinates and descriptors.
 
	\item DiffSAC can be seamlessly applied to other robust estimation methods as a plug-and-play module, replacing their minimum set sampling module.

\end{itemize}

\section{Limitations and Future Work}
\label{Limitations}
Despite the strong performance demonstrated, DiffSAC has several limitations that open avenues for future research.

On the one hand, the current framework requires task-specific models. A separate diffusion model must be trained for each distinct geometric estimation task, such as fundamental matrix estimation or homography. While effective, this approach lacks generality and requires a dedicated training process for every new problem type. A promising direction for future work is to explore the development of a single, more generalized model capable of handling multiple robust estimation tasks, which would significantly reduce training overhead and improve versatility.

On the other hand, there is a computational cost associated with the iterative nature of the diffusion model. While we achieve real-time performance on a GPU by leveraging accelerators like DPM-Solver++, the inference process can be demanding for resource-constrained environments. The reliance on a GPU may limit the applicability of DiffSAC on devices with only CPU capabilities or in embedded systems. Therefore, future research could focus on developing more lightweight and efficient diffusion samplers to reduce the computational burden without sacrificing performance.

\section{Conclusion}
This paper presents DiffSAC, a novel robust estimation method leveraging deep learning. To improve efficiency, DiffSAC integrates the established sample consensus process with a diffusion model. The diffusion model serves to identify and discard bad minimum sets early in the process. This allows for iterative refinement of confidence, leading to more reliable and precise estimations. DiffSAC then selects data points exhibiting high confidence to form minimum sets. Geometric features guide the generation of the diffusion model, ensuring relevant solutions. To further enhance the quality of the estimated model, DiffSAC generates and evaluates multiple high-quality minimum sets using sample consensus to identify the optimal hypothesis. The sampling approach allows DiffSAC to function as a plug-and-play module that is readily integrable into other robust estimation methods based on sample consensus. Experiments demonstrate that DiffSAC achieves state-of-the-art performance. Its design also facilitates straightforward application to diverse robust estimation tasks.

\backmatter

\bmhead{Acknowledgements}
This work was supported in part by the Natural Science Foundation of China under Grant 62225309, U24A20278, 62361166632 and U21A20480.

\bmhead{Data Availability}
Our implementation is available at \url{https://github.com/IRMVLab/DiffSAC}.

\bmhead{Competing interests}
The authors declare no competing interests.

\begin{appendices}

\section{Network Architecture Details}\label{secA1}
The network structures are based on Point-e \cite{nichol2022point}. To maintain permutation invariance, no positional encoding is applied.

In the attention feature extraction module, the features are first layer-normalized and then passed through a standard transformer network \cite{vaswani2017attention}. The output is layer-normalized again. Finally, a fully connected layer maps the features into a vector to denoise the confidence $c_t$ into $c_{t-1}$. The detailed network parameters are shown in Table \ref{tab:denoser}. For the MLP, a $1 \times 1$ convolution with a stride of 1 is used.

\begin{table*}[h!]\small 
\centering
\caption{\textbf{Detailed network parameters of the Confidence Denoise Neural Network.} ``Width" indicates the number of output channels.}
\vspace{-4pt}
\setlength{\tabcolsep}{2.0mm}
\renewcommand\arraystretch{1.3}
\begin{tabular}{cccc}
\toprule
Module                                     & \multicolumn{2}{c}{Layer}                                          & Parameter          \\ \hline
\multirow{7}{*}{Feature Embedding}      & \multirow{1}{*}{Confidence $c_t$}    & Fully Connected    & \emph{Width=$\left[512\right]$}    \\ \cmidrule{2-4} 
                                           & \multirow{3}{*}{$t$}    & MLP             & \emph{Width=$\left[512\right]$}               \\
                                           &                             & GELU                                 & - \\ 
                                            &                           & Repeating                 & - \\  \cmidrule{2-4} 
                                           & \multirow{2}{*}{Data points $\chi$}    & Repeating             & if inference               \\
                                           &                & Fully Connected           & \emph{Width=$\left[512\right]$} \\  \cmidrule{2-4}
                                           &     & Add    & - \\ \hline 
\multirow{5}{*}{Attention Feature Extraction}   & \multicolumn{2}{c}{LayerNorm}        & \emph{Width=$\left[512\right]$}      \\ \cmidrule{2-4}
                                        & & \\
                                        & \multicolumn{2}{c}{Transformer}        & \emph{Width=$\left[512\right]$}, \emph{Layers=$\left[12\right]$}, \emph{Heads=$\left[8\right]$}      \\ 
                                        & & \\ \cmidrule{2-4}
                                        & \multicolumn{2}{c}{LayerNorm}        & \emph{Width=$\left[512\right]$}      \\ \hline
Denoising              & \multicolumn{2}{c}{Fully Connected}                    & \emph{Width=$\left[512\right]$}      \\ \bottomrule
\end{tabular}
\label{tab:denoser}
\end{table*}

\end{appendices}

\end{document}